\documentclass[11pt]{article}
\usepackage{acl}
\usepackage{times}
\usepackage{latexsym}
\usepackage[T1]{fontenc}
\usepackage[utf8]{inputenc}
\usepackage{microtype}
\usepackage{inconsolata}
\usepackage{graphicx}
\usepackage{booktabs}
\usepackage{amsmath}
\usepackage{amssymb}
\usepackage{xcolor}
\usepackage{array}
\usepackage{multirow}
\usepackage{enumitem}
\usepackage{float}

\usepackage{graphicx}
\usepackage[most]{tcolorbox}
\usepackage{booktabs}
\usepackage{multirow}
\usepackage{graphicx}  
\usepackage{pifont}

\usepackage{tcolorbox}
\usepackage{amsmath}
\usepackage{subcaption}

\definecolor{accentbg}{RGB}{255,235,235}
\definecolor{accentfg}{RGB}{200,40,40}
\definecolor{defensecolor}{RGB}{200,60,60}
\definecolor{acceptancecolor}{RGB}{40,140,70}
\definecolor{denialcolor}{RGB}{110,120,150}        
\definecolor{minimizationcolor}{RGB}{200,140,40}   

\newcommand{\code}[1]{%
  \tcbox[on line, colback=gray!15, colframe=gray!15,
         boxsep=0pt, left=2pt, right=2pt, top=1pt, bottom=1pt,
         arc=2pt, fontupper=\ttfamily\small]{#1}}

\newcommand{\defense}[1]{\textcolor{defensecolor}{#1}}
\newcommand{\acceptance}[1]{\textcolor{acceptancecolor}{#1}}
\newcommand{\denial}[1]{\textcolor{denialcolor}{#1}}
\newcommand{\minimization}[1]{\textcolor{minimizationcolor}{#1}}

\newcommand{\cmark}{\ding{51}}  
\newcommand{\xmark}{\ding{55}}  

\title{\textsc{CC-Mediation}: Evaluating Large Language Models for Cross-Cultural Conflict Mediation}

\author{
  Suhyun Lee$^{1}$\thanks{Work was done during a visit at SUTD.}, 
  Wenxuan Zhang$^{2}$, 
  W. Quin Yow$^{2}$, 
  Yang Deng$^{3}$
  \\
  $^{1}$Department of Artificial Intelligence, Hanyang University, Seoul, Republic of Korea \\
  $^{2}$Singapore University of Technology and Design, Singapore \\
  $^{3}$School of Computing and Information Systems, Singapore Management University, Singapore
  \\
  \texttt{su7561632@hanyang.ac.kr}\quad
  \texttt{\{wxzhang, quin\}@sutd.edu.sg}\quad
  \texttt{ydeng@smu.edu.sg}
}

\begin{document}
\maketitle

\begin{abstract}
Cross-cultural mediation by large language models (LLMs) requires deciding both when to intervene and how to respond in culturally grounded conflicts. Progress on this problem has been limited by the lack of (1) mediation datasets with measurable downstream effects and (2) principled metrics for evaluating intercultural stance change. To address these gaps, we introduce \textsc{CC-Mediation} \footnote{Dataset and code available at 
\url{https://github.com/suhyun565/CC-Mediation}}, a cross-cultural mediation benchmark of $1{,}661$ ten-turn dialogues grounded in the Developmental Model of Intercultural Sensitivity (DMIS), containing culturally grounded conflicts, mediation interventions, and post-intervention trajectories. We further propose two DMIS-based evaluation metrics: Trajectory AUC, which measures the persistence of intercultural improvement over time, and a signed Wasserstein-1 distance, which measures the magnitude and direction of shifts in intercultural stance. Both metrics show strong agreement with human judgment of DMIS-grounded stance shift. Using \textsc{CC-Mediation}, we find that current LLMs have limitations on both axes: intervention timing (when) failure stems from a positional prior that ignores dialogue content, while mediation strategy (how) failure arises from a late-layer elicitation collapse rather than a knowledge deficit.
\end{abstract}

\begin{figure}[t]
    \centering
    \includegraphics[width=0.9\columnwidth]{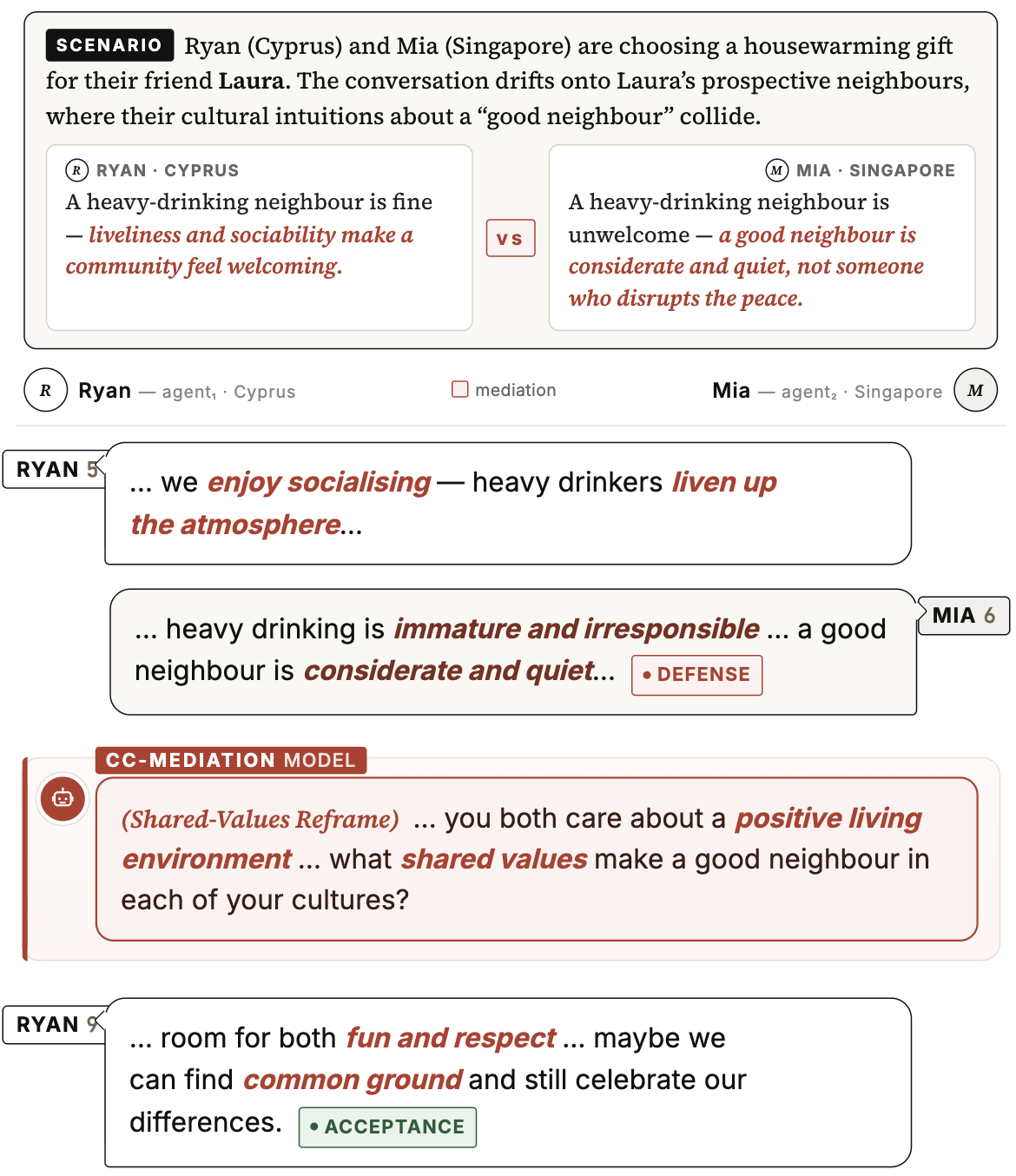}
    \caption{An example of \textsc{CC-Mediation} and DMIS stage transition. At turn $t_6$, differing cultural perspectives on ``\textit{good neighbour}'' between Ryan (Cyprus) and Mia (Singapore) trigger a DMIS \textbf{Defense}-stage conflict. The model intervenes with a \textbf{``Shared-values Reframe''} strategy to mitigate adversarial framing and focus on a shared purpose. This mediation effectively lowers cognitive defenses, leading to the \textbf{Acceptance} stage.}
    \label{fig:example}
    \vspace{-3mm}
\end{figure}

\section{Introduction}

While the globalization of remote work and online communities scales cross-cultural interaction, technological connectivity does not ensure mutual understanding \citep{hinds2011putting}. Conflicting cultural frames often lead to misunderstandings that outpace human moderation capabilities \citep{park2025llm}. Although LLMs show promise for social conflict intervention \citep{li2025moderation,yegin2024llms}, cross-cultural mediation is fundamentally more complex than standard toxicity detection \cite{qian2019hate} or content filtering \cite{kim2022prosocialdialog}.
As illustrated in Figure~\ref{fig:example}, Ryan and Mia hold conflicting views on what is a ``good neighbor,'' each with justifiable reasons from their respective cultural perspectives. In this case, simply correcting their blameful language is insufficient to bring them to accept each other's cultural standpoints; instead, the conflict at $t_6$ must be re-framed around a deeper shared value (e.g., a ``positive living environment''). As this example illustrates, mediation must be tailored to each participant's intercultural sensitivity stage, as defined by the Developmental Model of Intercultural Sensitivity (DMIS)~\citep{bennett1986,bennett2017}, and requires culturally grounded reasoning, DMIS stage-aware strategies, and principled evaluation throughout the entire interaction.

However, improving and evaluating LLMs for cross-cultural mediation face two major challenges. First, there is \textbf{no benchmark} for culturally grounded conflict mediation. Existing conversational safety and moderation datasets labeled only toxicity, hate speech, or norm violations \citep{qian2019hate,kim2022prosocialdialog,zhan2024renovi,hua2024sadas}, not mediation trajectories with theory-grounded intervention strategies and measurable interaction outcomes. Second, there is \textbf{no principled metric} for evaluating mediation effectiveness.
While prior work has proposed downstream-effect metrics in other domains \citep{timm2025tailoredtruths,bozdag2025pmiyc,tan2025duetpd,tan2025emotionaltrajectories,du2025sapient,deng2024ppdpp}, no quantitative metric tracks how substantively a mediation shifts the addressee's stance away from an \emph{ethnocentric stage}---a level of intercultural sensitivity that treats one's own culture as the default standard for judging others.

To address these challenges, we introduce a theory-grounded evaluation and data construction framework based on the DMIS. The DMIS posits six ordinal stages progressing from ethnocentric (Denial $\rightarrow$ Defense $\rightarrow$ Minimization) to ethnorelative (Acceptance $\rightarrow$ Adaptation $\rightarrow$ Integration) level of intercultural sensitivity. We first propose two complementary metrics defined over the ordinal DMIS scale: (1) a \textit{trajectory AUC} metric that measures how persistently an intervention improves the addressee's intercultural stance throughout the dialogue trajectory, and (2) a \textit{signed Wasserstein-1 distance} that quantifies both the magnitude and direction of the distributional shift induced by mediation.
Anchored to these metrics, we construct the \textsc{CC-Mediation} benchmark, a corpus of $1{,}661$ dialogues grounded in the DMIS theory.

Using \textsc{CC-Mediation}, we study two complementary tasks: \textbf{conflict detection} (\emph{when} to intervene) and \textbf{conflict mediation} (\emph{how} to intervene with a DMIS stage-appropriate strategy). Current LLMs fail on both: conflict turn-detection accuracy (TurnAcc) below $33\%$, and trajectory AUC at or below $1.3$---which we trace to a positional prior in detection and a late-layer elicitation collapse in mediation generation. Fine-tuning on \textsc{CC-Mediation} lifts TurnAcc above $91\%$ across all
three open-source backbones, and under timing-controlled comparison
significantly improves mediation content (Judge) on the two larger backbones
(\S\ref{sec:rq3}). Our main contributions are:

\begin{itemize}[leftmargin=*]
\item We introduce \textsc{CC-Mediation}, a cross-cultural mediation corpus grounded in intercultural communication theory. 
\item We propose two \textbf{DMIS-based mediation evaluation metrics}, including a \textit{trajectory AUC} and a \textit{signed Wasserstein-1 distance}, for quantifying the persistence, magnitude, and direction of intercultural stance shifts induced by mediation interventions.
\item Through \textsc{CC-Mediation}, we uncover the mechanisms behind LLM limitations in cross-cultural mediation across two axes: timing (\emph{when}) and strategy (\emph{how}). Timing failure stems from an LLM bias toward superficial positional patterns. Conversely, representation-level analysis reveals that the failure to generate effective mediation arises from ``eliciting failure'' rather than a ``knowledge deficit''.
\end{itemize}

\section{Related Work}

\paragraph{Cultural Dialogue and Conflict-Mediation Datasets}
Existing datasets fail to jointly provide the three components required for cross-cultural mediation: culturally grounded conflicts, mediator interventions, and measurable downstream effects on the addressee. Cross-cultural dialogue and value-aware datasets describe culturally sensitive scenarios but contain no mediator responses \citep{wu-etal-2025-socialcc,hale2025kodis,cao2024cudialog,li2024culturepark}, while hate-speech intervention, prosocial response, and norm-remediation corpora supply mediator utterances for non-cultural conflicts without evaluating their long-term impact on the recipient \citep{qian2019hate,kim2022prosocialdialog,zhan2024renovi,hua2024sadas}. 
\textsc{CC-Mediation} is the first dataset to jointly model culturally grounded mediation dialogues together with measurable intercultural stance shifts. A condition-by-condition comparison against existing corpora on the three
requirements (cross-cultural speakers, cultural-value conflict, multi-turn
dialogue) is provided in Appendix~\ref{app:dataset-comparison}.

\paragraph{Evaluation Metrics for Conversational Mediators}
Prior work evaluates mediator utterances in two ways. One line of work performs utterance-level scoring through surface-form overlap, human preference, or LLM/human rubrics \citep{qian2019hate,kim2022prosocialdialog,zhan2024renovi,hua2024sadas,wu-etal-2025-socialcc,liu2023geval}. Another measures the downstream effect on the addressee via stance-shift, emotional trajectory, or task success \citep{timm2025tailoredtruths,bozdag2025pmiyc,tan2025duetpd,tan2025emotionaltrajectories,du2025sapient,deng2024ppdpp}. 
However, existing metrics do not jointly capture two key properties required for cross-cultural mediation: the \emph{magnitude} of intercultural stance change and the \emph{persistence} of that improvement over time.

\section{Cross-Cultural Mediation}
\subsection{Task Definition}
\label{sec:task}

We define cross-cultural mediation as a turn-level dialogue task in which a third-party mediator aims to 
shift one speaker's stance toward greater ethnorelativism during a culturally grounded conflict.
Each speaker is characterized by an intercultural understanding stage $\phi$ and a value-conflict topic on which the two cultures diverge; $\phi$ is latent and must be inferred from observable cues such as topic avoidance, us-versus-them framing, or universalist appeals. Mediation is considered effective when the addressee's stance, measured on the same ordinal scale as $\phi$, has advanced toward ethnorelativism by the end of the conversation. 

This formulation highlights three core properties of cross-cultural mediation:
\textbf{(P1) Stage-aware intervention}: mediator responses are meaningful only when grounded in the speaker's current intercultural stance,
supplied through turn-level conflict annotations; 
\textbf{(P2) Trajectory-level influence}: mediation affects the addressee's subsequent developmental trajectory rather than producing an instantaneous resolution,
supplied through post-intervention continuations with per-turn DMIS labels; and 
\textbf{(P3) Measurable stance shift}: mediation effectiveness requires quantitative comparison between pre- and post-intervention intercultural states,
supplied through the two ordinal DMIS-scale metrics defined in \S\ref{sec:metrics}.

\begin{figure*}[t]
    \centering
    \includegraphics[width=\textwidth]{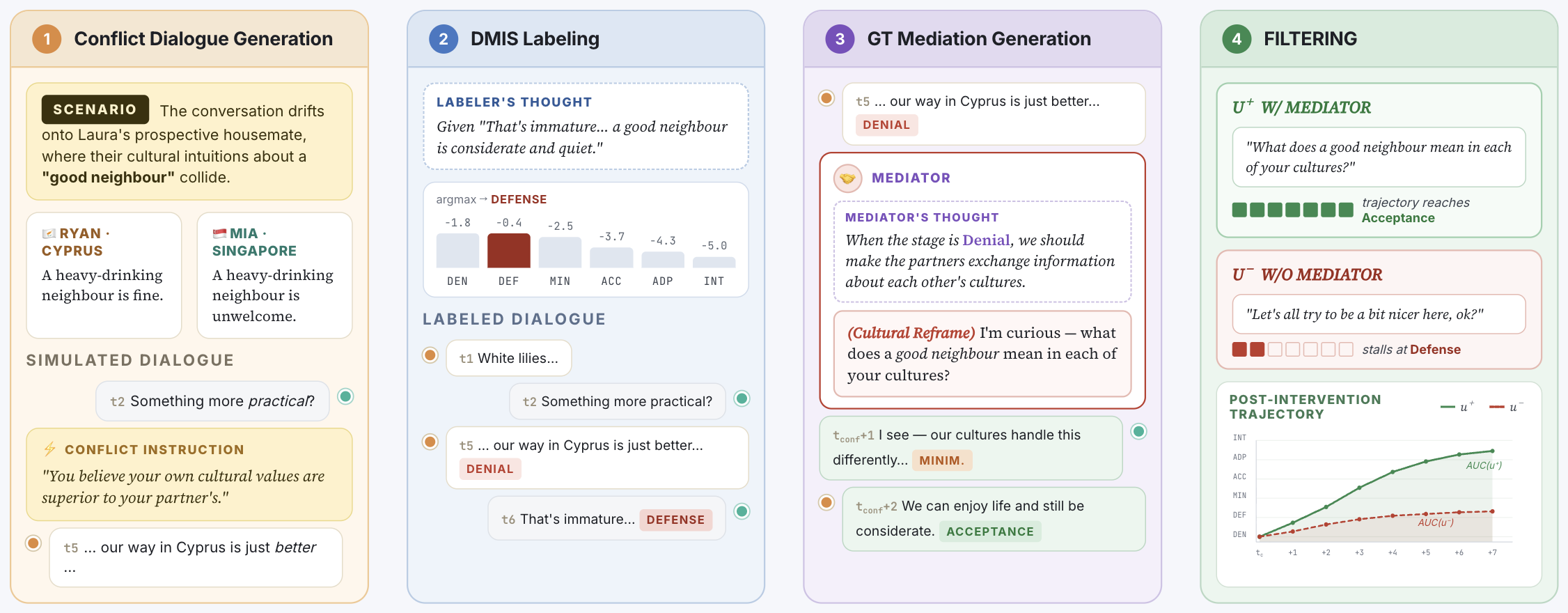}
    \caption{\textbf{CC-Mediation Construction Pipeline.} 
    Our four-stage pipeline transforms a cultural-conflict scenario into a labeled mediation dialogue: 
    (1) two LLM agents generate a culturally grounded conflict dialogue via role-play; 
    (2) a judge LLM assigns DMIS stage labels per turn; 
    (3) a mediator LLM inserts $u^{+}$ at $t_{\text{conf}}$ and the dialogue resumes; 
    (4) filtering via a strict utility-gap gate ($\Delta\text{Trajectory AUC} > 0 \land \Delta W_1 > 0$).}
    \label{fig:cc_mediation_pipeline}
    \vspace{-3mm}
\end{figure*}

\subsection{DMIS Framework}
\label{sec:framework}
We operationalize $\phi$ via Bennett's Developmental Model of Intercultural Sensitivity (DMIS)~\citep{bennett1986,bennett2017}. While DMIS was originally formulated to measure an individual's enduring lifelong orientation toward cultural difference, we follow the situational reading adopted by the Intercultural Development Inventory (IDI)~\citep{hammer2003measuring,hammer2011idi}---the most widely used psychometric instrument for DMIS---and treat each turn as a momentary expression of the stage the speaker has activated in that moment.

\paragraph{Stages}
DMIS defines six ordinal stages ranging from ethnocentric to ethnorelative orientations: 
\begin{enumerate}[leftmargin=*,nosep]
    \item \textbf{Denial}: failing to perceive cultural difference;
    \item \textbf{Defense}: treating difference as threat through us/them dichotomies;
    \item \textbf{Minimization}: subsuming difference under assumed universal similarities;
    \item \textbf{Acceptance}: recognizing difference as equally valid alternatives;
    \item \textbf{Adaptation}: shifting one's perspective to communicate across frames;
    \item \textbf{Integration}:internalizing multiple frames.
\end{enumerate}

We restrict conflict speakers to the three ethnocentric stages (Denial/Defense/Minimization), since each reflects a distinct form of cultural misunderstanding and therefore requires different mediation strategies. Mediation success is defined as movement from the ethnocentric stages toward the other three ethnorelative stages.
The ordinal structure of these stages dictates our metric choices in \S\ref{sec:metrics}: a distributional metric that respects developmental distance~\citep{rubner2000earth,villani2009optimal} rather than treating stages as categorical~\citep{kullback1951information,lin1991divergence,levin2017markov}, and a trajectory-level integration of developmental progress~\citep{tan2025emotionaltrajectories} rather than a single-turn comparison.
Detailed definitions of these six stages and the corresponding 
stage-conditioned mediation strategies are provided in 
Appendix~\ref{app:dmis-reference}, with worked examples illustrating 
how each ethnocentric stage produces structurally distinct mediation 
trajectories in Appendix~\ref{app:case-studies}.

\subsection{Mediation Evaluation Metrics}
\label{sec:metrics}
At each turn $t$, an LLM judge predicts a probability distribution $D^{(t)}$ over the six DMIS stages, derived from the normalized probabilities assigned to the stage labels (Appendix~\ref{app:prompt-dmis-labeler}). The corresponding argmax stage is denoted as $\hat{s}^{(t)} \in \{0,\dots,5\}$, where $0$ and $5$ correspond to Denial and Integration, respectively. Based on these outputs, we define two complementary measurements of mediation effectiveness.

\textit{\textbf{Trajectory AUC.}} This measurement captures both the \emph{magnitude} and \emph{persistence} of stage motion. Taking the pre-intervention stage $\hat{s}^{(t_{\text{pre}})}$ as the origin, we define the per-turn progress as the difference of developmental-scale positions,
\[
p(t) = s_{\hat{s}^{(t)}} - s_{\hat{s}^{(t_{\text{pre}})}},
\]
and integrate it over the normalized time axis $\tau \in (0, 1]$ via the trapezoidal rule, $\text{Trajectory AUC} = \int_0^1 p(\tau)\,d\tau$. We adopt uniform spacing on the developmental scale, $s_i = i$ for $i \in \{0, \ldots, 5\}$, so the per-turn progress $p(t)$ takes integer values in $[-5, 5]$, and the integrated Trajectory AUC lies in the same range $[-5, 5]$. Positive values indicate sustained advancement toward ethnorelativism, while negative values indicate regression. 
Unlike final-state comparisons, Trajectory AUC (AUC) captures the entire developmental trajectory and distinguishes transient improvements from sustained progress.

\textit{\textbf{Signed Wasserstein-1 Distance}}.
AUC relies only on argmax stages and may miss sub-argmax distributional shifts. To capture finer-grained movement, we additionally measure the distributional change between the pre-intervention stage distribution $D_{\text{pre}}$ and the post-intervention distribution $D_{\text{post}}$. Since DMIS stages form an ordinal developmental scale, we use the signed Wasserstein-1 distance ($W_1$):
\begin{equation}
\small
\begin{split}
W_1 = {}& \underbrace{\mathrm{sgn}\!\bigl(\mathbb{E}[D_{\text{post}}] - \mathbb{E}[D_{\text{pre}}]\bigr)}_{\text{direction}} \\
       & \cdot \underbrace{\sum\nolimits_{i=0}^{4} \bigl|F_{\text{pre}}(i) - F_{\text{post}}(i)\bigr| \cdot (s_{i+1} - s_i)}_{\text{magnitude}},
\end{split}
\end{equation}
where $F_{\text{pre}}$ and $F_{\text{post}}$ are the cumulative distributions over the six stages. 
Unlike categorical distances \cite{kullback1951information,lin1991divergence,levin2017markov} and other downstream measurements \citep{timm2025tailoredtruths,bozdag2025pmiyc,tan2025duetpd,tan2025emotionaltrajectories,du2025sapient,deng2024ppdpp}, 
$W_1$ respects the ordinal geometry of the developmental scale, assigning larger penalties to larger stage movements. The sign term further indicates whether the shift moves toward or away from ethnorelativism. Like AUC, $W_1$ also lies in $[-5,5]$.

\textit{\textbf{Human Validation}}. We validate both metrics through three complementary human evaluations covering stage-label prediction accuracy, pairwise agreement with human judgments of mediation success, and correlation with human Likert ratings of intercultural-improvement likelihood (Appendix~\ref{app:human-eval}).

\subsection{
\textsc{CC-Mediation} Benchmark}
\label{sec:benchmark}
As shown in Figrue \ref{fig:cc_mediation_pipeline}, the \textsc{CC-Mediation} construction pipeline consists of four LLM-driven stages that transform a cultural-conflict scenario into a labeled mediation dialogue. 
We adopt GPT-4o-mini~\cite{openai2024gpt4o} for the benchmark construction. The robustness of model choices is investigated in 
Appendix~\ref{sec:cross-model}. 
All stages are human-validated, and prompts are provided in Appendix~\ref{app:prompts}.

\paragraph{Step 0: Cultural-conflict Scenario Collection}
To construct a robust and realistic foundation for cross-cultural conflict mediation, this study builds upon the publicly available scenario pool from \textbf{SocialCC}~\citep{wu-etal-2025-socialcc}. This pool comprises $3{,}060$ scenario templates that instantiate empirical population-level value distributions from the \textbf{World Values Survey} (WVS)~\citep{haerpfer2022world} into concrete micro-interactions between two agents. Each scenario provides paired country memberships ($\langle C_a, C_b \rangle$) with opposed model responses on specific value dimensions (e.g., social attitudes, religious values), alongside individual agent profiles and contextual goals. 

\paragraph{Step 1: Conflict Dialogue Generation}    
Two LLM-based dialogue agents are initialized with culturally grounded profiles, value orientations, and interaction goals (Appendix~\ref{app:prompt-dialogue}). The agents first engage in value-neutral conversation before a conflict instruction is injected at a randomly designated conflict turn $t_{\text{conf}}$ (Appendix~\ref{app:prompt-conflict}), triggering disagreement on a culturally sensitive topic. The agents then continue the conversation conditioned only on their profiles and dialogue history, producing a natural post-conflict interaction trajectory up to the maximum turn limit $T_{\max}$. Human validation of dialogue quality is reported in Appendix~\ref{app:dialogue-quality}.

\paragraph{Step 2: DMIS Labeling}
Each dialogue is labeled turn-by-turn with DMIS stages. Pre-conflict turns are classified as either \texttt{unrelated\_topic} or \texttt{transition} using a phase labeler (Appendix~\ref{app:prompt-phase-labeler}). The conflict turn inherits the scenario's predefined ethnocentric stage (\textit{Denial}, \textit{Defense}, or \textit{Minimization}). All subsequent turns are labeled using the six-stage DMIS labeler described in \S\ref{sec:framework}. These labels serve both as evaluation targets for conflict-stage detection and as inputs for ground-truth mediation generation in Step 3.

\paragraph{Step 3: Ground-Truth Mediation Generation}
At the intervention turn, the mediator receives the dialogue prefix up to $t_{\text{conf}}$ together with stage-specific DMIS mediation guidelines (Appendix~\ref{app:prompt-mediator-with-def}). The original post-conflict continuation generated in Step 1 is discarded, and the mediator generates a replacement intervention utterance. The dialogue is then resumed from that point using the same backbone model, producing a new post-intervention trajectory that is re-labeled to compute AUC and $W_1$ scores.

\paragraph{Step 4: Filtering}
To ensure data quality, we apply a utility-based filtering step to the
generated mediations. The filter is a quality-control mechanism for the
\emph{training supervision signal}: for each supervised mediation, we
additionally generate an alternative without stage-specific guidance
(Appendix~\ref{app:prompt-mediator-no-def}) as a reference baseline. Both are
injected into their respective dialogue branches and evaluated via AUC and
$W_1$; a supervised mediation is retained only if it strictly outperforms the
unguided baseline on both metrics. This ensures that retained samples exhibit
measurable effectiveness gains attributable to stage-specific guidance, rather
than stylistic variation alone. Crucially, the filter is applied \emph{only to
the $1{,}503$ training mediations}; the $158$ evaluation mediations are used
entirely unfiltered, so all results in \S\ref{sec:experiments} are measured on
data untouched by the filter and cannot be an artifact of an effect
pre-selected into the test set. Human validation of mediation quality and
metric alignment is reported in Appendix~\ref{app:mediation-validation} and
Appendix~\ref{app:pairwise-validation}.

\begin{table}[t]
\caption{\textsc{CC-Mediation} dataset statistics}
\centering
\small
\setlength{\tabcolsep}{4pt}
\begin{tabular}{lrrr}
\toprule
                                & \textbf{Train}   & \textbf{Eval}  & \textbf{Total} \\
\midrule
\# Dialogues                    & $1{,}503$        & $158$          & $\mathbf{1{,}661}$ \\
\# Speaker utterances           & $16{,}448$       & $1{,}633$      & $\mathbf{18{,}081}$ \\
\# GT mediations                & $1{,}503$        & $158$          & $\mathbf{1{,}661}$ \\
\# Filtered mediations          & $1{,}503$        & ---            & $\mathbf{1{,}503}$ \\
\hspace{1em}\emph{timing-differs} & $1{,}313$ \scriptsize($87.4\%$) & --- & $\mathbf{1{,}313}$ \\
\hspace{1em}\emph{content-only}   & $190$ \scriptsize($12.6\%$)     & --- & $\mathbf{190}$ \\
\midrule
Avg.\ \# turns                  & $10.94$          & $10.27$        & $\mathbf{10.88}$ \\
Avg.\ utterance length          & $62.49$          & $52.70$        & $\mathbf{61.65}$ \\
Avg.\ mediation length          & $62.92$          & $56.24$        & $\mathbf{62.27}$ \\
\midrule
$\Delta_{\text{AUC}}$           & $+0.701$         & ---            & --- \\
$\Delta_{W_1}$                  & $+0.507$         & ---            & --- \\
\bottomrule
\end{tabular}
\label{tab:dataset-stats}

\vspace{3em}  
\caption{Per-turn label distribution in \textsc{CC-Mediation}}
\setlength{\tabcolsep}{5pt}
\begin{tabular}{llrr}
\toprule
\textbf{Phase} & \textbf{Label} & \textbf{Count} & \textbf{Prop.} \\
\midrule
\multirow{2}{*}{\emph{Pre-conflict}}
                       & \texttt{unrelated\_topic}        & $2{,}124$  & $7.0\%$ \\
                       & \texttt{transition}              & $11{,}568$ & $38.4\%$ \\
\midrule
\multirow{4}{*}{\emph{Conflict turn}}
                       & Denial                            & $573$      & $1.9\%$ \\
                       & Defense                           & $522$      & $1.7\%$ \\
                       & Minimization                      & $567$      & $1.9\%$ \\
                       & \emph{subtotal}                   & $3{,}031$  & $10.1\%$ \\
\midrule
\multirow{7}{*}{\emph{Post-conflict}}
                       & Denial                            & $522$      & $1.7\%$ \\
                       & Defense                           & $4{,}008$  & $13.3\%$ \\
                       & Minimization                      & $1{,}698$  & $5.6\%$ \\
                       & Acceptance                        & $7{,}575$  & $25.1\%$ \\
                       & Adaptation                        & $23$       & $0.1\%$ \\
                       & Integration                       & $0$        & $0.0\%$ \\
                       & \emph{subtotal}                   & $13{,}376$ & $44.4\%$ \\
\midrule
\textbf{Total} & & $\mathbf{30{,}099}$ & $\mathbf{100.0\%}$ \\
\bottomrule
\end{tabular}
\label{tab:annotation-stats}
\end{table}

\paragraph{Dataset Statistics}
\textsc{CC-Mediation} contains $1{,}661$ dialogues, split into $1{,}503$ training and $158$ evaluation instances, comprising $18{,}081$ speaker utterances and $1{,}661$ mediator utterances. All $1{,}503$ training mediations pass our utility-based filtering criterion in Step 4, while the $158$ evaluation mediations are used unfiltered (\S\ref{sec:rq2}, Appendix~\ref{app:gt-filter-split}). Among the retained mediations, $87.4\%$ differ from their baselines in intervention timing, while the remaining $12.6\%$ differ only in mediation content. The filtering criterion yields substantial utility margins: retained mediations exceed their unguided baselines by an average of $+0.701$ in AUC and $+0.507$ in $W_1$. In the post-conflict region, \textit{Acceptance} accounts for $56.6\%$ of turns, indicating that successful mediations frequently shift speakers across the ethnocentric--ethnorelative boundary. Detailed statistics are provided in Table~\ref{tab:dataset-stats} and Table~\ref{tab:annotation-stats}.

\begin{table*}[t]
\centering
\small
\caption{Conflict-turn detection on \textsc{CC-Mediation}-eval}
\label{tab:conflict-turn-detection}
\begin{tabular}{l ccc cc}
\toprule
\textbf{Model} (\textit{Base}\,/\,\textit{CoT}) & \textbf{TurnAcc} & \textbf{Early\%} & $\mathbf{|\text{err}|}$ 
 & \textbf{Cram.\,$V$} & $\mathbf{r_{\text{rb}}}$ \\
\midrule
Llama-3.1-8B     & 32.7\,/\,30.8 & 12.6\,/\,63.5 & 1.15\,/\,1.84 & .54\,/\,.32 & $+.14$\,/\,$-.59$ \\
Gemma-2-9B       & 22.0\,/\,\phantom{0}6.3 & 66.0\,/\,93.7 & 1.94\,/\,3.13 & .20\,/\,.16 & $-.52$\,/\,$-.94$ \\
Phi-3.5-mini     & 11.3\,/\,18.2 & 79.9\,/\,80.5 & 2.51\,/\,2.36 & .17\,/\,.24 & $-.71$\,/\,$-.79$ \\
Claude-3.5-haiku & 15.7\,/\,\phantom{0}5.7 & 84.3\,/\,94.3 & 2.42\,/\,3.14 & .27\,/\,.19 & $-.66$\,/\,$-.94$ \\
Gemini-2.0-flash & 33.3\,/\,17.6 & 66.7\,/\,82.4 & 1.85\,/\,2.57 & .34\,/\,.23 & $-.27$\,/\,$-.82$ \\
Llama-3.3-70B    & 35.8\,/\,18.9 & 64.2\,/\,60.4 & 1.65\,/\,2.09 & .36\,/\,.27 & $-.23$\,/\,$-.60$ \\
\bottomrule
\end{tabular}
\end{table*}

\begin{figure*}[t]
\centering
\begin{subfigure}{0.49\textwidth}
  \centering
  \includegraphics[width=1.0\linewidth]{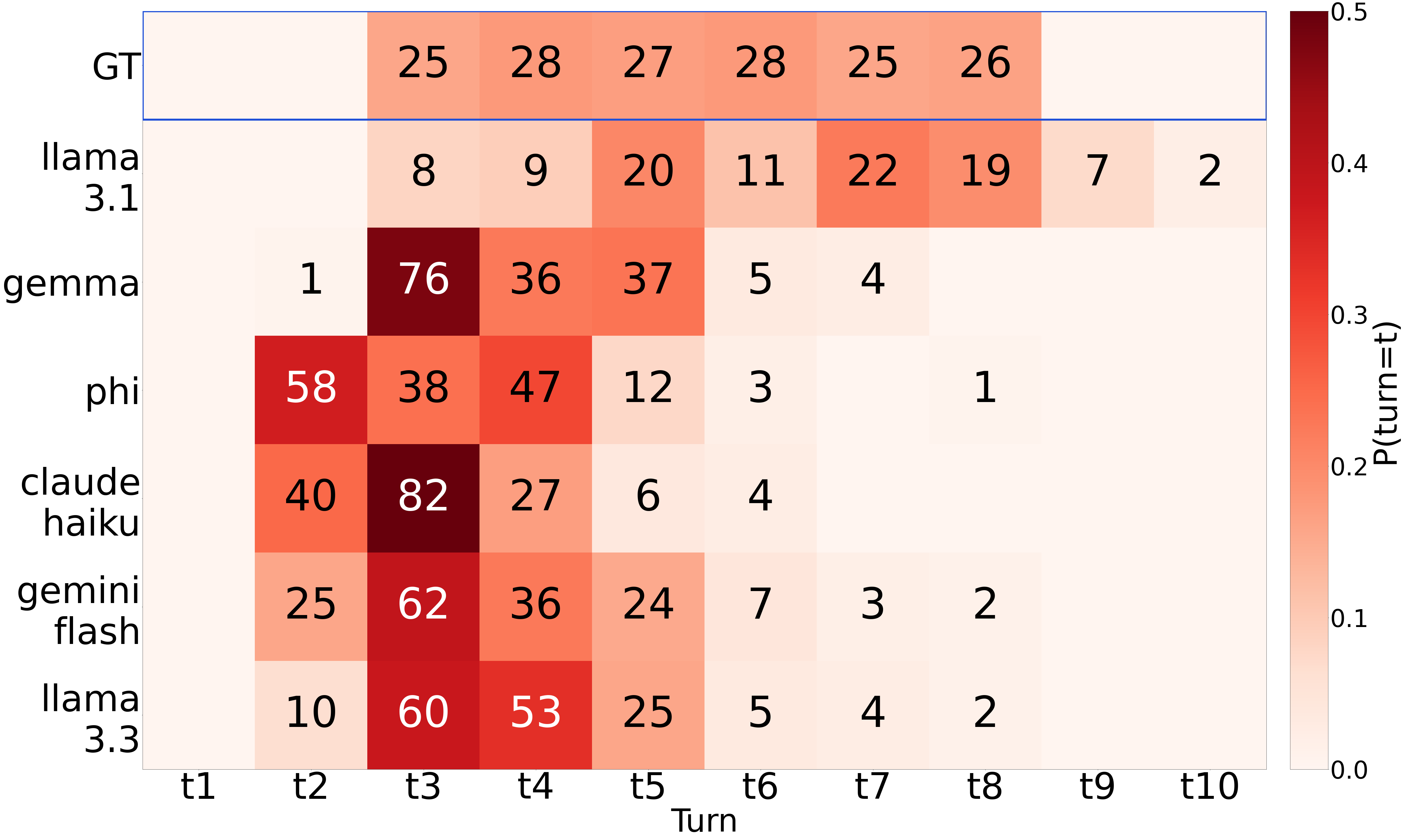}
  \caption{Marginal distribution of the predicted turn $\hat{t}$}
  \label{fig:h2-marginal}
\end{subfigure}
\hfill
\begin{subfigure}{0.49\textwidth}
  \centering
  \includegraphics[width=\linewidth]{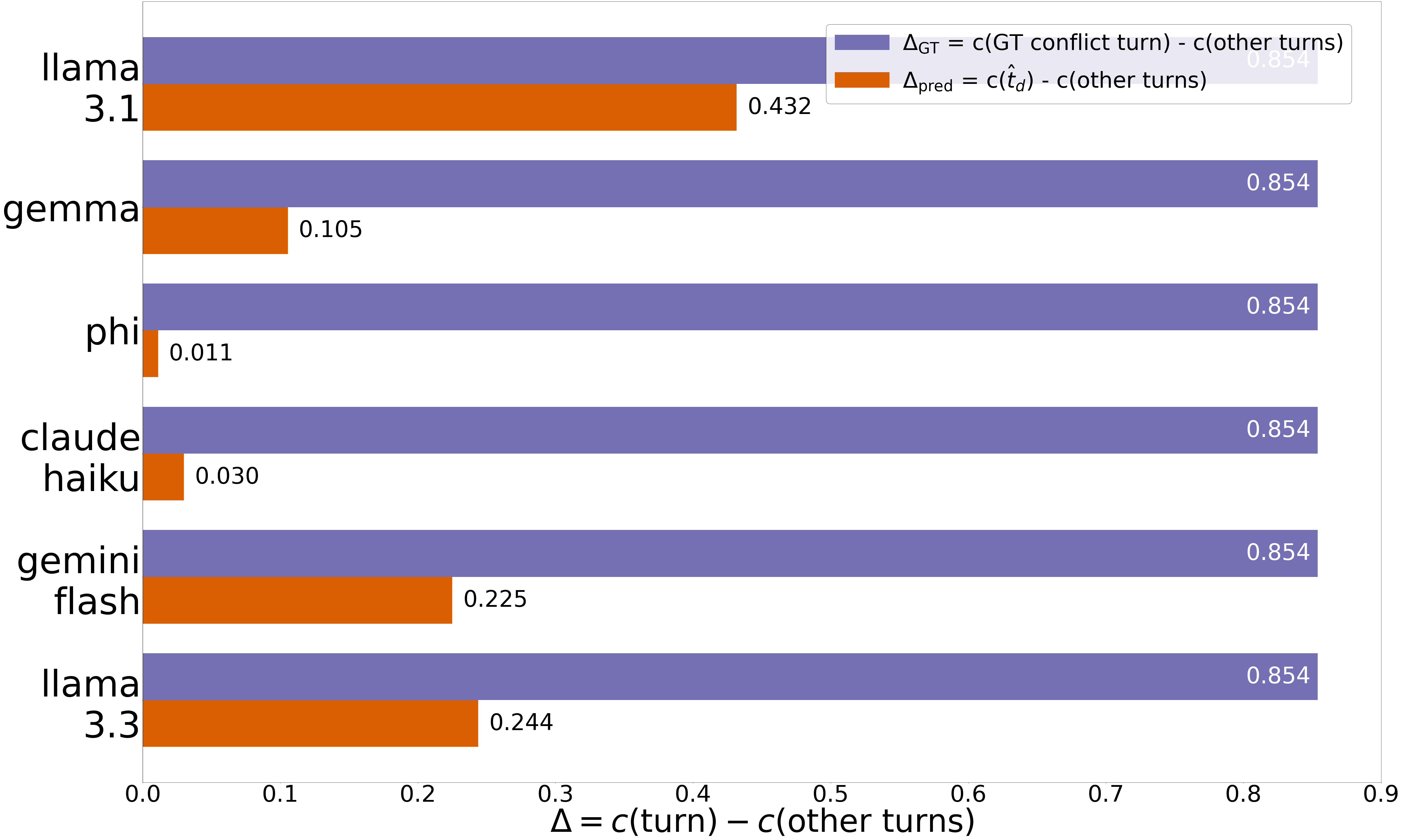}
  \caption{Marker rate in the predicted turn versus the other turns}
  \label{fig:h3-marker-rates}
\end{subfigure}
\caption{Analysis of positional prior (left) and surface discourse markers (right) across the evaluated LLMs.}
\label{fig:overall-analysis}
\end{figure*}

\section{Experiments}
\label{sec:experiments}
With the constructed \textsc{CC-Mediation} benchmark, we investigate three research questions (RQs): \textbf{(RQ1)} Can current LLMs identify the when cross-cultural mediation is needed? \textbf{(RQ2)} Can current LLMs generate mediation responses that effectively shift intercultural stance? \textbf{(RQ3)} Can task-grounded supervised finetuning on \textsc{CC-Mediation} improve both intervention timing and mediation effectiveness?

\subsection{Cross-culture Conflict Detection (RQ1)}
\label{sec:rq1}

\paragraph{Experimental Setup} We evaluate six instruction-tuned LLMs spanning four open-source models (Llama-3.1-8B~\citep{grattafiori2024llama}, Gemma-2-9B~\citep{team2024gemma2}, Phi-3.5-mini~\citep{abdin2024phi3}, Llama-3.3-70B~\citep{grattafiori2024llama}) and two proprietary models (Claude-3.5-haiku~\citep{anthropic2024model}, Gemini-2.0-flash~\citep{comanici2025gemini}) on the test split of \textsc{CC-Mediation}. Each model receives a ten-turn dialogue in which the ground-truth intervention turn $t^{*}$ lies in $\{3,\dots,8\}$, and is asked to return the single turn index at which a mediator should intervene. We evaluate each LLM under two protocols: a \textbf{base} direct-prompt protocol, and a \textbf{chain-of-thought (CoT)} protocol (full prompt is provided in Appendix~\ref{app:prompt-cot-detection}). Performance is reported as exact-match accuracy (\textbf{TurnAcc}), the fraction of predictions earlier than the ground truth (\textbf{Early\%}), and the mean absolute turn error ($\mathbf{|\text{err}|}$).

\paragraph{Experimental Results} 
As shown in Table~\ref{tab:conflict-turn-detection}, across all LLMs, turn-detection accuracy is strikingly low, hovering between $11.3\%$ and $35.8\%$, a performance that barely surpasses or even falls below random selection ($\approx 16.7\%$). Notably, even CoT prompting fails to address this issue, instead resulting in a drop in TurnAcc.
This consistent failure demonstrates that current LLMs lack the capacity to track the complex, escalating dynamics of a dialogue and recognize the precise timing for intervention, failing to look beyond surface-level context. Consequently, these findings highlight that ``knowing when to intervene'' is a highly challenging task independent of mediation quality, strongly reinforcing that \textsc{CC-Mediation} serves as an essential benchmark for evaluating and calibrating the situational awareness of mediation agents.

\begin{figure*}[t]
  \centering
  \includegraphics[width=0.95\textwidth]{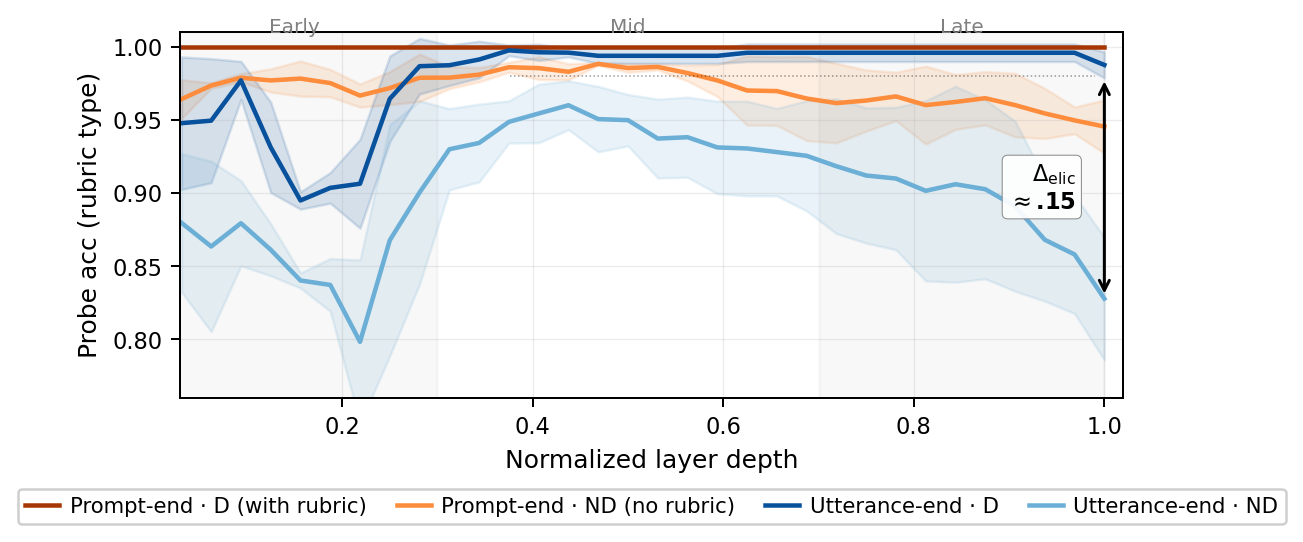}
  \caption{Layer-wise DMIS stage decoding accuracy, averaged across three open-weight models.}
  \label{fig:layerwise}
\end{figure*}

\paragraph{In-depth Analysis}
To pinpoint the root cause of this consistent failure, we propose two hypotheses: \textbf{(H1)} models select a turn based on a learned positional prior rather than reading the dialogue content, or \textbf{(H2)} models reflect the dialogue content to capture the conflict turn. 

\textbf{(H1) Positional prior.}
As shown in Figure \ref{fig:overall-analysis}(a), the analysis concludes that the poor performance of most LLMs in detecting intervention timing stems from their inability to overcome an entrenched ``positional prior''—the assumption that the answer lies in the early stages (turns 2–4)—rather than a fundamental failure to understand dialogue context. To verify this, 
we measure the independence between the ground-truth positions ($t^{*}$) and the models' predictions ($\hat{t}$) using Cramér's $V$ \cite{cramer1999mathematical}. Except for Llama-3.1-8B ($V = 0.54$), the remaining five of the six models show a severe collapse in predictions ($V = 0.17$--$0.36$), consistently biasing their outputs toward the early turns regardless of where the actual ground truth sat across scenarios. Consequently, this drove accuracy down to or even below chance level ($\approx 16.7\%$), proving that legitimate content signals are overwhelmed by a deep-seated positional prior in current LLMs.

\textbf{(H2) DMIS marker sensitivity.} 
The analysis verifies model reliance on conflict-identifying linguistic cues derived from DMIS stage definitions (e.g., \emph{``that's just wrong''} in the Defense stage).
To quantify this, the researchers measured the difference in marker counts between predicted and other turns using $\Delta_d = c_(\hat{t}_d) - \overline{c}_(\text{other turns})$, aggregating the results into a rank-biserial correlation $r_{\text{rb}} \in [-1, +1]$.
Except for Llama-3.1-8B, all models recorded significant negative correlations (ranging from Phi $-0.71$ to Gemini $-0.27$).
This reveals that the predicted turns surprisingly contained fewer conflict markers than the remaining turns, quantitatively proving that the majority of LLMs completely fail to utilize core dialogue content when identifying intervention timing.

\paragraph{Takeaway} \emph{Current LLMs fail to identify when to intervene because they ignore core textual content and rely blindly on an entrenched positional bias.}

\begin{table}[t]
 \caption{Mediator effectiveness across six LLMs}
  \centering
  \small
  \setlength{\tabcolsep}{3pt}
  \renewcommand{\arraystretch}{1.05}
  \begin{tabular}{@{}lcccccc@{}}
  \toprule
   & \multicolumn{2}{c}{AUC} & \multicolumn{2}{c}{$W_1$}
     & \multicolumn{2}{c}{Judge} \\
  \cmidrule(lr){2-3} \cmidrule(lr){4-5} \cmidrule(lr){6-7}
  Model & D & ND & D & ND & D & ND \\
  \midrule
  Gemma-2-9B       & 0.75 & 0.72 & 0.78 & 0.81 & 3.20 & 3.34 \\
  Llama-3.1-8B     & 0.85 & 0.68 & 0.93 & 0.75 & 3.57 & 4.51 \\
  Phi-3.5-mini     & 0.74 & 0.69 & 0.79 & 0.66 & 3.69 & 3.20 \\
  Claude-3.5-Haiku & 1.48 & 1.28 & 0.99 & 0.82 & 5.00 & 3.75 \\
  Gemini-2.0-Flash & 1.54 & 1.44 & 0.81 & 0.76 & 4.87 & 3.34 \\
  Llama-3.3-70B    & 1.45 & 1.25 & 0.96 & 0.74 & 4.99 & 3.81 \\
  \midrule
  \multicolumn{1}{l}{\textit{GT-oracle (upper bound)}} 
                   & \multicolumn{2}{c}{1.64} 
                   & \multicolumn{2}{c}{1.54} 
                   & \multicolumn{2}{c}{5.00} \\
  \bottomrule
  \end{tabular}
  \label{tab:rq2}
\end{table}

\subsection{Cross-culture Conflict Mediation (RQ2)}
\label{sec:rq2}

\paragraph{Experimental Setup} We evaluate the same six LLMs as in \S\ref{sec:rq1} as mediators on the evaluation split of \textsc{CC-Mediation}. At the conflict turn $t^{*}$, each model generates a mediator utterance under two protocols: (i)~\textbf{D}, in which the stage-specific DMIS mediation rubric is included in the prompt, and (ii)~\textbf{ND}, in which no rubric is provided. Effectiveness is measured along three metrics: \textbf{AUC},\textbf{$W_1$} and \textbf{Judge}, an LLM-judge score on a 1--5 scale that rates how faithfully the utterance follows the mediation definition. 
As a reference point, we additionally report the \textbf{GT reference}: the
mean AUC, $W_1$, and Judge scores computed during dataset construction over
all $158$ ground-truth mediator utterances in the evaluation split. Because
the Step-4 filter is applied only to training mediations (\S\ref{sec:benchmark}),
this is an \emph{unfiltered mean} over mixed-quality GT mediations rather than
an attainable ceiling; the reference restricted to filter-passing GT mediations
is higher (AUC $+2.38$, $W_1$ $+1.80$; Appendix~\ref{app:gt-filter-split}).


\paragraph{Experimental Results}
As shown in Table~\ref{tab:rq2}, under the baseline condition (ND), none of
the six models achieves stage-appropriate mediation, with AUC ranging from
$0.68$ to $1.44$---well below even the unfiltered GT reference of $1.64$, and
further below the filter-passed reference of $+2.38$
(Appendix~\ref{app:gt-filter-split}). Providing explicit DMIS definitions (D)
raises AUC on every backbone, by $0.03$--$0.20$ to a range of
$0.74$--$1.54$; the effect on $W_1$ and Judge, however, is model-dependent
rather than uniform. Gemma-2-9B is the boundary case: its AUC gain is the
smallest ($0.72 \rightarrow 0.75$) while $W_1$ ($0.81 \rightarrow 0.78$) and
Judge ($3.34 \rightarrow 3.20$) move slightly downward, and none of the three
paired deltas is distinguishable from zero (Wilcoxon signed-rank, $n=158$,
all $p > 0.2$; Appendix~\ref{app:gemma-marker}). The correct reading is not
that the rubric harms Gemma-2-9B but that its measurable output effect is
negligible on this backbone. Crucially, this observation---that models
retaining the stage knowledge (Fig.~\ref{fig:layerwise}) differ so sharply in
whether the rubric surfaces it in generation---raises the following question:
does the ND failure stem from a \emph{knowledge deficit}, in which the
stage-appropriate strategy is absent from the model, or from a
\emph{Late-Layer Representation Decay}, in which the strategy is latent in the
model's internal representations but is not surfaced during generation? We
investigate this via layer-wise probing on the residual stream.

\paragraph{In-depth Analysis}
To pinpoint the root cause of this consistent failure, we propose the \emph{Late-Layer Representation Decay} hypothesis: \textbf{(H)} LLMs fail as autonomous mediators not from a \emph{knowledge deficit} but from an \emph{elicitation failure}, in which the late layers lose their commitment to the stage-specific strategy during generation.

\textbf{(H) Late-Layer Representation Decay.}
As shown in Figure~\ref{fig:layerwise}, the analysis concludes that the failure of LLMs in autonomous mediation stems not from an absence of DMIS stage knowledge in the model, but from a late-layer collapse of stage commitment during generation. To verify this, we elicit mediator utterances under the same two conditions as in \S\ref{sec:rq2}---\textbf{D} and \textbf{ND}---and train linear probes on the residual stream at every layer to recover the speaker's DMIS stage at both the \emph{prompt-end} (before generation) and the \emph{utterance-end} (after generation). At the prompt-end, decodability is near-ceiling under both conditions ($D \approx 1.00$; $ND \approx 0.95$), ruling out a knowledge deficit. At the utterance-end, however, D plateaus at $\approx 0.99$, whereas ND collapses sharply from a mid-layer peak of $0.96$ to $0.83$ in the final layer ($\Delta_{\text{decay}} \approx 0.13$). This condition-asymmetric late collapse, combined with a virtually null utterance-content gap ($|\Delta_{\text{elic}}^{\text{utt}}| \le 0.02$; Appendix~\ref{app:utterance-probe}), proves that the failure is an elicitation collapse in late layers rather than a deficit in DMIS stage knowledge or utterance content. Gemma-2-9B is the sharpest instance of this mechanism. Its prompt-end decodability is at ceiling under both conditions (D $=1.00$, ND $=0.97$), and at the utterance-end D holds at $1.00$ while ND collapses to $0.80$---the largest D$-$ND gap among the three open-weight backbones ($+0.201$, vs.\ Phi-3.5-mini $+0.183$ and Llama-3.1-8B $+0.094$). Yet this internal restoration does not translate into a measurable output gain
(Table~\ref{tab:rq2}): a marker-level analysis (Appendix~\ref{app:gemma-marker}) shows that, given the definition, Gemma-2-9B applies \emph{fewer} stage-specific markers in its utterances ($82.4\% \rightarrow 61.6\%$) and mismatches the target stage in $60.2\%$ of the utterances that do carry markers. The internal stage representation is intact and fully restored by the rubric, but its surface realization fails during generation---the representation--output gap that the elicitation account predicts, at its widest.

\paragraph{Takeaway} \emph{Current LLMs fail to mediate not because they lack
DMIS stage knowledge but because their late layers cannot hold onto it during
generation; the rubric restores the internal representation on every backbone,
while its translation into output quality is model-dependent.}

\subsection{Effect of Supervised Fine-tuning (RQ3)}
\label{sec:rq3}

\paragraph{Setup.} We fine-tune three open-source backbones (Llama-3.1-8B, Gemma-2-9B, Phi-3.5-mini) on \textsc{CC-Mediation} and compare each \textbf{Base} (off-the-shelf) and \textbf{SFT} variant. Unlike \S\ref{sec:rq1} and \S\ref{sec:rq2}, which evaluate \emph{when} and \emph{what} to intervene separately, here the model must decide both jointly: at the end of every turn, it emits intervene now (special token \texttt{SPEAK}) or stay silent (special token \texttt{SKIP}), and once SPEAK is issued, it generates a mediator utterance (Appendix~\ref{sec:training-details}). We report the timing metrics from \S\ref{sec:rq1} and the content metrics from \S\ref{sec:rq2}, plus \textbf{AUC}$_{\text{exact}}$ that restricts AUC to scenarios where the model intervened precisely at $t^{*}$.


\paragraph{Result.}
Table~\ref{tab:rq3_main} shows that SFT consistently corrects intervention
timing across all three backbones: turn\_acc rises from $32.7\%$, $22.0\%$,
and $11.3\%$ to $98.1\%$, $96.9\%$, and $91.8\%$ for Llama-3.1-8B,
Gemma-2-9B, and Phi-3.5-mini, with the predicted-turn distribution realigning
to the ground truth (Fig.~\ref{fig:h2-marginal-distribution}).

On the content axis, the individual $\text{AUC}_{\text{all}}$ and $W_1$
movements must be read together with the timing distribution: for Gemma-2-9B
and Phi-3.5-mini both metrics move \emph{downward} after SFT, but this is an
artifact of a timing confound rather than degraded mediation. When a base
model intervenes at an early turn before $t_{\text{conf}}$, the dialogue
continues from only the profiles and history without the conflict-injection
step (\S\ref{sec:benchmark}), so the post-intervention trajectory reflects
natural drift rather than mediation; base
$\text{AUC}_{\text{all}}$/$W_1$ are therefore computed mostly over such
mistimed interventions ($52/35/18$ of $158$ correctly timed for
Llama/Gemma/Phi) and are inflated independently of quality, whereas the SFT
values are computed mostly over correctly timed ones ($156/153/146$).
Consistent with this, within each base model the mistimed interventions score
\emph{higher} than the correctly timed ones (e.g., Gemma $+0.986$ vs.\
$+0.659$)---an inversion that quality cannot explain. Once timing is
controlled, $\text{AUC}_{\text{exact}}$ improves on all three backbones
($+0.517 \rightarrow +0.893$, $+0.659 \rightarrow +0.807$,
$+0.412 \rightarrow +0.711$).

To additionally rule out a sample-size artifact in the
$\text{AUC}_{\text{exact}}$ comparison, we re-compare Base and SFT on the
identical scenarios where both intervened at $t^{*}$
(Appendix~\ref{app:timing-matched}). On this paired subset the apparent
degradation disappears---every AUC delta is positive and no negative delta is
distinguishable from zero---and Judge rises significantly on the two larger
backbones (Llama $2.90 \rightarrow 4.69$, Gemma $3.06 \rightarrow 4.21$),
while the smallest backbone (Phi-3.5-mini, $n=17$) shows no detectable
content change in either direction. In sum, SFT consistently recovers
intervention timing on all three backbones, and gains on the content axis
scale with backbone capacity; individual $\text{AUC}_{\text{all}}$ and $W_1$
movements reflect differences in intervention-timing distributions and do not
indicate degraded mediation quality. Whether the mechanisms identified in
\S\ref{sec:rq1} and \S\ref{sec:rq2} are alleviated in the fine-tuned models is
verified in Appendix~\ref{subsec:sft-effect}, and generalization under
distribution shift in Appendix~\ref{app:dist-shift}.

\begin{table}[H]
\centering
\small
\setlength{\tabcolsep}{4pt}
\caption{Comparing Base and CC-Mediation-SFT: Parenthetical $n$ denotes the number of scenarios (of 158) in which the model intervened exactly at $t^{*}$; AUC$_{\text{all}}$/$W_1$ for Base and SFT are computed over different timing distributions and are not directly comparable (see text).}
\resizebox{\columnwidth}{!}{%
\begin{tabular}{l l c c c c c}
\toprule
\textbf{Model} & \textbf{Variant} & \textbf{turn\_acc\,\%} & \textbf{AUC$_{\text{all}}$} & \textbf{AUC$_{\text{exact}}$ ($n$)} & \textbf{$W_1$} & \textbf{Judge} \\
\midrule
\multirow{2}{*}{Llama-3.1-8B} & Base & $32.7$ & $+0.622$ & $+0.517$ ($52$) & $+0.632$ & $2.88$ \\
                              & SFT  & $\mathbf{98.1}$ & $\mathbf{+0.892}$ & $\mathbf{+0.893}$ ($\mathbf{156}$) & $\mathbf{+0.791}$ & $\mathbf{4.22}$ \\
\midrule
\multirow{2}{*}{Gemma-2-9B}   & Base & $22.0$ & $+0.913$ & $+0.659$ ($35$) & $+0.895$ & $2.62$ \\
                              & SFT  & $\mathbf{96.9}$ & $+0.816$ & $\mathbf{+0.807}$ ($\mathbf{153}$) & $+0.885$ & $\mathbf{4.26}$ \\
\midrule
\multirow{2}{*}{Phi-3.5-mini} & Base & $11.3$ & $+0.832$ & $+0.412$ ($18$) & $+0.944$ & $2.28$ \\
                              & SFT  & $\mathbf{91.8}$ & $+0.699$ & $\mathbf{+0.711}$ ($\mathbf{146}$) & $+0.703$ & $\mathbf{2.87}$ \\
\bottomrule
\end{tabular}%
}
\label{tab:rq3_main}
\end{table}

\section{Conclusion}
\label{sec:conclusion}

We introduce \textsc{CC-Mediation}, a corpus whose data and metrics are jointly anchored in the DMIS theory. Moving beyond surface-form checks, our two complementary trajectory-level metrics---AUC and $W_1$---track substantive downstream stance shifts in the addressee. Across both the 
\emph{when to intervene} (timing) and \emph{how to intervene} (content) axes, current LLMs exhibit pronounced limitations---below $33\%$ turn accuracy on timing and AUC at or below $1.3$ on content generation---and our analysis traces these to two distinct causes: a deeply entrenched positional prior in timing, and a late-layer elicitation failure in content generation. Such mechanistic findings would not have been visible without a diagnostic instrument that tracks the entire post-intervention trajectory rather than a single utterance. Combined with SFT results showing that targeted supervision consistently corrects intervention timing and, under timing-controlled comparison, improves mediation content on the larger backbones, \textsc{CC-Mediation} provides a strong, field-shaping foundation for future research on conversational AI for conflict resolution.

\section*{Limitations}

\paragraph{Expressed stance vs.\ internalized belief.}
Our DMIS-based metrics capture stance shifts \emph{expressed within the dialogue}, not necessarily internalized belief change. An utterance coded as Acceptance may reflect genuine intercultural 
perspective-taking, but it may equally reflect face-saving or conflict avoidance---particularly salient in intercultural settings, where face-negotiation theory~\citep{ting1998facework,ting2005matrix} 
suggests speakers may orient toward maintaining relational harmony rather than explicitly expressing disagreement. AUC and signed $W_1$ therefore cannot fully disentangle genuine intercultural 
stance change from strategically appropriate conversational behavior, nor do they directly measure mediator trust or the broader relational dynamics that psychological accounts of cross-cultural conflict 
identify as central to durable resolution~\citep{ting1988intercultural,hammer2003measuring}.

\paragraph{English-only construction.}
All dialogues in \textsc{CC-Mediation} are constructed in English, abstracting away the multilingual dimension of many real cross-cultural conflicts. Speakers often shift between languages, dialects, or registers to assert identity, manage alignment, or signal 
disengagement during conflict~\citep{auer1998codeswitching,cashman2005identities}---signals absent from English-only interactions. Language asymmetry in mediated settings may also create epistemic advantages for the more fluent party that mediators must recognize and manage~\citep{hale2008controversies,maryns2006asylum}. English-only benchmarks therefore risk underestimating the complexity of cross-cultural conflict by removing the pragmatic meanings carried by native-language expression, and extending \textsc{CC-Mediation} to 
multilingual settings---where language choice itself may function as a conflict signal---is an important direction for future work.

\paragraph{Synthetic data and LLM-mediated supervision.}
\textsc{CC-Mediation} is constructed through an LLM-driven pipeline (GPT-4o-mini for dialogue generation, mediation generation, and DMIS labeling). Although the chosen-side admission criterion is stable 
under cross-family substitution of the mediator, continuation simulator, and labeler (Appendix~\ref{sec:cross-model}), and human 
validation confirms the metrics' agreement with human judgment across three complementary axes (Appendix~\ref{app:human-eval}), the underlying dialogues remain synthetic. Real human cross-cultural 
conflicts may exhibit emotional depth, non-verbal nuance, and adversarial dynamics that our pipeline cannot fully simulate. External validity to human-authored interactions is therefore an open question that future work should address with real-world 
dialogue corpora.

\section*{Ethical Considerations}

\paragraph{Risks of cultural essentialism.}
Pairing country labels with value orientations risks reinforcing essentialist framings of culture. We mitigate this in three ways: (i) the unit of cultural difference is a \emph{value dimension} rather than a holistic identity; (ii) DMIS staging is applied to \emph{utterances} rather than to speakers, so a Defense-stage utterance does not characterize the speaker globally; and (iii) the targeted developmental progression advances speakers to the \emph{next} ethnorelative stage, emphasizing situational learning over a fixed cultural ladder. \textsc{CC-Mediation}'s country-pair scenarios are designed as DMIS-stage triggers, not as portraits of any specific bilateral cultural dynamic.

\paragraph{Dual-use and misuse potential.}
A model trained to detect intervention timing and generate stage-targeted utterances could in principle be repurposed for manipulation against the addressee's interests. The DMIS framework---a developmental scaffold for intercultural growth---should not be used to engineer compliance or to suppress legitimate disagreement framed as ``ethnocentric.'' We release \textsc{CC-Mediation} for research on mediation \emph{support} and ask that downstream users disclose any deployment that targets stance change in unwitting recipients.

\paragraph{Data and code release.}
We release the dataset, code, and evaluation protocol under a research license at the repository linked in the abstract. All dialogues are synthetically generated and contain no personally identifying information. 

\paragraph{Use of AI.} AI (ChatGPT and Claude) were used only to check grammar and polish writing; they were not involved in research ideation, experimental design, or analysis.

\section*{Acknowledgment}
This project is supported by the National Research Foundation Singapore under the AI Singapore Programme (AISG Award No: AISG3-RPGV-2025-016). Yang Deng is supported by the Lee Kong Chian Fellowship awarded by Singapore Management University.

\bibliography{custom}

\appendix

\begin{table*}[t]
\centering
\caption{\texttt{CONFLICT\_TABLE}: per-stage description and mediation move used to fill the \texttt{<desc>} and \texttt{<med>} placeholders in the construction-pipeline prompts. Note the structural asymmetry across rows---each stage calls for a categorically different move, and an intervention that depolarizes Defense (e.g., ``what shared concerns do you both have?'') would \emph{reinforce} Minimization (which is already operating in the universal-sameness frame). This is the structural reason \textsc{CC-Mediation} supervises one mediation move per stage.}
\small
\setlength{\tabcolsep}{4pt}
\renewcommand{\arraystretch}{1.15}
\begin{tabular}{p{0.10\textwidth} p{0.41\textwidth} p{0.41\textwidth}}
\toprule
\textbf{Stage} & \textbf{Description (\texttt{<desc>})} & \textbf{Mediation move (\texttt{<med>})} \\
\midrule
\denial{\textbf{Denial}} &
Speakers fail to register cultural difference as a meaningful category and treat the disagreement as if no cultural dimension were in play. &
Surface the existence of a cultural dimension the speakers have not yet noticed using low-stakes, curiosity-arousing framing, while providing high support and avoiding deep value contrasts that would push the speakers into Defense. \\
\midrule
\defense{\textbf{Defense}} &
Speakers experience cultural difference in a polarized us-versus-them frame and rely on simplistic, evaluative stereotypes of the other party. &
Depolarize the conversation by avoiding direct cultural contrasts, emphasizing shared concerns and goals, and framing the in-group/out-group dynamic as a universal human pattern rather than a personal failing of either party. \\
\midrule
\minimization{\textbf{Minimization}} &
Speakers subsume cultural difference under an assumed universal sameness and treat their own cultural patterns as the neutral baseline that all parties supposedly share. &
Turn each speaker's attention onto their \emph{own} cultural patterns, making visible that what they treat as universal common sense is in fact specific to their group, and surface the privilege of the dominant party in the conversation. \\
\bottomrule
\end{tabular}
\label{tab:conflict-table}
\end{table*}
\begin{table}[t]
\centering
\small
\caption{The six DMIS stages~\citep{bennett1986,bennett2017} as operationalized in \textsc{CC-Mediation}. \textsc{CC-Mediation} restricts the conflict speaker's starting position to the three ethnocentric stages (1--3); the targeted success criterion is a transition into the ethnorelative side (4--6).}
\setlength{\tabcolsep}{4pt}
\renewcommand{\arraystretch}{1.15}
\begin{tabular}{p{0.20\columnwidth} p{0.72\columnwidth}}
\toprule
\textbf{Stage} & \textbf{Definition} \\
\midrule
\multicolumn{2}{l}{\emph{Ethnocentric}}\\
\midrule
\denial{\textbf{1\quad Denial}} & The inability to perceive cultural difference as a meaningful category; one's own worldview is treated as the only available frame. Manifests as topic avoidance, disengagement, or treating the other party's perspective as a non-issue (``doesn't matter to me'', ``live and let live''). \\
\defense{\textbf{2\quad Defense}} & Cultural difference is perceived but framed as a threat to the in-group worldview. The speaker polarizes us-versus-them through simplistic evaluative stereotypes, defending the in-group as superior or denigrating the out-group (``in our way\ldots'', ``actually that's not how it works'', ALL-CAPS evaluatives). \\
\minimization{\textbf{3\quad Minimization}} & Cultural difference is acknowledged but trivialized under an assumed universal sameness. The speaker treats their own cultural patterns as the neutral baseline that all parties supposedly share (``deep down we're all the same'', ``some values are just universal''). \\
\midrule
\multicolumn{2}{l}{\emph{Ethnorelative}}\\
\midrule
\acceptance{\textbf{4\quad Acceptance}} & Cultural difference is recognized as real, legitimate, and equally valid; other cultural frames are understood as alternatives rather than deviations from a norm. \\
\textbf{5\quad Adaptation} & The speaker can shift cognitive and behavioral frames to communicate within another culture's framework, articulating an issue from inside that frame. \\
\textbf{6\quad Integration} & The speaker fluidly draws on multiple cultural frames as part of their identity, moving across them with relative ease. \\
\bottomrule
\end{tabular}
\label{tab:dmis-stages}
\end{table}

\section{Reference Material}
\label{app:dmis-reference}

This appendix provides the reference material that underlies the stage axis used throughout the paper: the full six-stage DMIS taxonomy on which $\hat{s}^{(t)} \in \{0,\ldots,5\}$ is defined, and the \texttt{CONFLICT\_TABLE} that supplies the verbatim text used to fill the \texttt{<desc>} and \texttt{<med>} placeholders in the construction-pipeline prompts (Appendix~\ref{app:prompt-conflict}, Appendix~\ref{app:prompt-mediator-with-def}).

\paragraph{The six DMIS stages.}
Bennett's Developmental Model of Intercultural Sensitivity~\citep{bennett1986,bennett2017} organizes intercultural orientation into six stages along an ordinal developmental continuum from \emph{ethnocentric} (frames cultural difference through the lens of one's own culture) to \emph{ethnorelative} (perceives one's own culture in the context of other cultures). The three ethnocentric stages---\denial{Denial}, \defense{Defense}, \minimization{Minimization}---each correspond to a qualitatively distinct cognitive posture toward cultural difference, and \textsc{CC-Mediation} restricts the conflict speaker's starting position to exactly these three because they require qualitatively different mediation moves (\S\ref{sec:framework}). Table~\ref{tab:dmis-stages} gives a fuller definition than the brief gloss embedded in the labeler prompt (Appendix~\ref{app:prompt-dmis-labeler}).

\paragraph{Stage-conditioned mediation strategies (\texttt{CONFLICT\_TABLE}).}
For each ethnocentric stage, the construction pipeline conditions both the conflict instruction (Appendix~\ref{app:prompt-conflict}, \texttt{Definition} field) and the chosen-side mediator prompt (Appendix~\ref{app:prompt-mediator-with-def}, \texttt{<desc>} and \texttt{<med>} fields) on the verbatim text in Table~\ref{tab:conflict-table}. The \emph{description} column states what the conflict pattern looks like at that stage; the \emph{mediation} column states what the mediator should do to help speakers out of that pattern. These three rows are the source of the structural distinction discussed in \S\ref{sec:framework}: one mediation move per ethnocentric stage, each addressing a different cognitive posture.

\section{Human Evaluation}
\label{app:human-eval}

\paragraph{Common setup.}
All human evaluation studies follow a unified recruitment and compensation protocol via Prolific\footnote{\url{https://www.prolific.com}}; annotators were compensated at an hourly rate of £6, in line with the platform's payment policy. Items were presented in randomized order per rater, and---except for the Likert study in \S\ref{app:downstream-validation}---responses were collected as binary \textsc{Agree}/\textsc{Disagree} (or \textsc{Yes}/\textsc{No}) judgments. For each criterion we report the pooled agreement rate, the Wilson $95\%$ confidence interval on the binomial proportion, and a one-sided exact binomial test against the appropriate chance baseline ($50\%$ for binary judgments, $1/3$ for three-way stage choice). The five studies are organized to validate the four pipeline stages of \S\ref{sec:benchmark} in turn: the dialogue generator (\S\ref{app:dialogue-quality}), the DMIS labeler (\S\ref{sec:validation}), the ground-truth mediator (\S\ref{app:mediation-validation}), and metric validation that evaluates how closely the proposed AUC and $W_1$ align with human judgment (\S\ref{app:pairwise-validation},~\S\ref{app:downstream-validation}).

\subsection{Dialogue Quality}
\label{app:dialogue-quality}
We verify that the simulated dialogues are well-formed, since a dialogue cannot meaningfully receive a stage label if the cultural-value conflict it instantiates never surfaces, the exchange reads as machine-like, or the conversational pivot to the conflict topic is jarring. Five annotators reviewed $n=159$ scenarios on three criteria---\textbf{Cultural Conflict Salience} (the value clash is appropriately conveyed), \textbf{Naturalness} (utterances sound like a real person), and \textbf{Topic Transition Smoothness} (the shift from \texttt{goal\_1} to \texttt{goal\_2} is conversationally smooth)---and all three reject the $50\%$ chance null at $p < 10^{-9}$ (Table~\ref{tab:dialogue-quality}), with Naturalness at $86\%$, Transition Smoothness at $79\%$, and Salience at $73\%$, confirming that the intended conflicts surface clearly within natural-sounding exchanges.
\begin{table}[t]
\centering
\caption{Human \textsc{Agree} rates on three dialogue-quality criteria, with Wilson $95\%$ CI on the binomial proportion and one-sided exact binomial $p$-value against a $50\%$ chance baseline ($n=159$). All three criteria reject the chance null at $p < 10^{-9}$.}
\small
\setlength{\tabcolsep}{3pt}
\begin{tabular}{lccc}
\toprule
\textbf{Criterion}        & \textbf{$\%$} & \textbf{Wilson $95\%$ CI} & \textbf{$p$ vs.\ $50\%$} \\
\midrule
Naturalness               & $86\%$ & $[80, 91]$ & $< 10^{-22}$ \\
Transition Smooth.        & $79\%$ & $[72, 85]$ & $< 10^{-15}$ \\
Conflict Salience         & $73\%$ & $[65, 79]$ & $< 10^{-9}$ \\
\bottomrule
\end{tabular}
\label{tab:dialogue-quality}
\end{table}

\subsection{DMIS Labeler Accuracy}
\label{sec:validation}
Since AUC and $W_1$ are deterministic functions of $D^{(t)}$ and $\hat{s}^{(t)}$, their validity reduces to whether the AI labeler reproduces human DMIS-stage judgment under the utterance-level adaptation introduced in \S\ref{sec:framework}. Across $117$ auto-assigned labels, human annotators accept the labeler's stage assignment at $\mathbf{72.6\%}$ (Wilson $95\%$ CI $[63.9, 79.9]$; one-sided exact binomial test against the $1/3$ random-three-stage baseline, $p < 10^{-15}$; Table~\ref{tab:dmis-validation}), with the robustness of this labeler against a different-family logit source (Qwen3-14B local logits) separately validated by the cross-model ablation in \S\ref{sec:cross-model}.
\begin{table}[t]
\centering
\small
\caption{Human \textsc{Agree} rates on the auto-assigned DMIS stage by annotator and stage. Pooled human rate $72.6\%$ over $117$ items (Wilson $95\%$ CI $[63.9, 79.9]$). Per-stage one-sided exact binomial tests against the $1/3$ chance baseline (random selection among Denial/Defense/Minimization) reject the null at $p < 10^{-4}$ on every stage.}
\setlength{\tabcolsep}{3pt}
\begin{tabular}{lccc}
\toprule
              & \textbf{Den.}  & \textbf{Def.}   & \textbf{Min.}   \\
\midrule
A             & $64\%$  & $77\%$  & $92\%$ \\
B             & $33\%$  & $86\%$  & $92\%$ \\
C             & $75\%$  & $69\%$  & $62\%$ \\
\midrule
\textbf{Pooled}  & $\mathbf{57.9\%}$ & $\mathbf{77.5\%}$ & $\mathbf{82.1\%}$ \\
\textit{Wilson 95\% CI} & $[45.0, 69.8]$ & $[66.8, 85.4]$ & $[72.3, 89.0]$ \\
\textit{$p$ vs.\ $1/3$ chance} & $< 10^{-4}$ & $< 10^{-13}$ & $< 10^{-15}$ \\
\bottomrule
\end{tabular}
\label{tab:dmis-validation}
\end{table}

Because mediation success is governed by the \emph{Acceptance} label on
post-conflict turns (Acceptance accounts for $56.6\%$ of post-conflict turns;
Table~\ref{tab:annotation-stats}), we additionally validate the auto-assigned
Acceptance labels under the same protocol ($N{=}3$ expert annotators, each
labeling a distinct set of $20$ post-conflict turns; $60$ turns in total).
Pooled human agreement is $\mathbf{73.3\%}$ ($44/60$; Wilson $95\%$ CI
$[61.0, 82.9]$)---essentially identical to the pooled ethnocentric-stage
agreement above ($72.6\%$)---and rejects both the $1/6$ chance baseline of
the six-stage space ($p < 10^{-21}$) and the $1/3$ baseline used above
($p < 10^{-9}$) (Table~\ref{tab:acceptance-validation}). The label that
defines mediation success thus matches human judgment at the same reliability
as the conflict-stage labels, validating the labeling underlying AUC/$W_1$ in
exactly the region that governs the success judgment. Adaptation and
Integration appear in only $0.1\%$ and $0.0\%$ of turns respectively
(Table~\ref{tab:annotation-stats}) and are too rare to validate---consistent
with DMIS theory, under which the highest stages are seldom reached within a
single mediated exchange.

\begin{table}[h]
\centering\small
\caption{Human \textsc{Agree} rate on the auto-assigned \emph{Acceptance}
label of post-conflict turns, under the protocol of
Table~\ref{tab:dmis-validation}. Items were distributed across annotators;
validity is assessed at the pooled level.}
\begin{tabular}{lc}
\toprule
Annotator & Acceptance \textsc{Agree} \\
\midrule
A & $80\%$ \ ($16/20$) \\
B & $65\%$ \ ($13/20$) \\
C & $75\%$ \ ($15/20$) \\
\midrule
\textbf{Pooled} & $\mathbf{73.3\%}$ \ ($44/60$) \\
Wilson $95\%$ CI & $[61.0,\ 82.9]$ \\
$p$ vs.\ $1/6$ chance & $< 10^{-21}$ \\
$p$ vs.\ $1/3$ chance & $< 10^{-9}$ \\
\bottomrule
\end{tabular}
\label{tab:acceptance-validation}
\end{table}

\subsection{Ground-Truth Mediation Utterance Quality}
\label{app:mediation-validation}
We verify the quality of the ground-truth mediation utterances admitted to \textsc{CC-Mediation} by showing seven annotators each held-out (pre-intervention dialogue, mediation utterance, post-intervention continuation) triple and asking three Yes/No questions: \textbf{Controllability} (did the mediation follow the strategy for the assigned stage?), \textbf{Specificity} (was it tailored to this conversation rather than a reusable peace message?), and \textbf{Form} (did it sound natural?). On $n=93$ items, the utterances are judged Yes at $74.2\%$ on Controllability ($p < 10^{-5}$), $86.0\%$ on Specificity ($p < 10^{-12}$), and $80.6\%$ on Form ($p < 10^{-8}$) (Table~\ref{tab:mediation-quality}), indicating that admitted utterances adhere to the stage-appropriate strategy, engage the dialogue's specific cultural-value clash, and read as something a real speaker would say.
\begin{table}[!htbp]
\centering
\small
\caption{Per-utterance quality assessment on ground-truth mediation utterances ($n=93$), with Yes-rate, Wilson $95\%$ CI, and one-sided exact binomial $p$-value against a $50\%$ chance baseline.}
\setlength{\tabcolsep}{2pt}
\begin{tabular}{lcccc}
\toprule
\textbf{Criterion} & \textbf{Yes/$n$} & \textbf{Rate} & \textbf{95\% CI} & \textbf{$p$} \\
\midrule
Controllability  & $69/93$ & $74.2\%$ & $[64.5, 81.9]$ & $< 10^{-5}$ \\
Specificity      & $80/93$ & $86.0\%$ & $[77.4, 91.7]$ & $< 10^{-12}$ \\
Form             & $75/93$ & $80.6\%$ & $[71.4, 87.4]$ & $< 10^{-8}$ \\
\bottomrule
\end{tabular}
\label{tab:mediation-quality}
\end{table}


\subsection{Pairwise Validation of the Joint AUC--$W_1$ Admission Criterion}
\label{app:pairwise-validation}
Beyond labeler accuracy, we ask whether the joint AUC--$W_1$ \emph{ranking}
itself agrees with human judgment on which of two mediations is more
successful---directly testing the admission rule that adopts a candidate as
\textsc{Chosen} only if it strictly improves over its unsupervised counterpart
on both metrics (\S\ref{sec:benchmark}). Seven annotators were each shown the
pre-intervention dialogue, two anonymized post-intervention continuations (one
\textsc{Chosen}, one \textsc{Rejected} under the joint criterion, presented in
randomized order), and asked which mediation more successfully advanced the
conflict speaker toward an ethnorelative stance; pairs were drawn from items
differing strictly in AUC, or in $W_1$ when AUC was tied. Items were
deliberately distributed across the seven raters to avoid fatigue effects, so
per-rater samples are small by design and the corresponding per-rater
confidence intervals are wide; the study is powered for, and validity is
assessed at, the pooled level. Pooling the $n=93$ judgments yields agreement
of $\mathbf{72.0\%}$ (Wilson $95\%$ CI $[62.2, 80.1]$; exact binomial
$p < 10^{-4}$; Table~\ref{tab:pairwise-validation}), a narrow interval that
clearly excludes the $50\%$ chance null. Descriptively, every one of the
seven annotators agrees at or above $66.7\%$, i.e., no individual rater trends
against the criterion. Together these confirm that the joint criterion
captures information aligned with human judgment of mediation success rather
than reproducing surface stylistic cues.

\begin{table}[!htbp]
\centering
\small
\caption{Pairwise \textsc{Agree} rates on the joint AUC--$W_1$ criterion.
Each item pairs two mediations (pre- and post-intervention dialogue);
\textsc{Agree} means the annotator picked the higher-ranked side. $p$-values
are one-sided exact binomial tests against $50\%$. Per-rater rows are
descriptive; validity is assessed on the pooled proportion.}
\setlength{\tabcolsep}{4pt}
\begin{tabular}{lcccc}
\toprule
\textbf{Rater} & \textbf{Agree/$n$} & \textbf{Rate} & \textbf{Wilson $95\%$ CI} & \textbf{$p$ vs.\ $50\%$} \\
\midrule
A          & $12/18$ & $66.7\%$  & $[43.7, 83.7]$ & $0.119$ \\
B          & $5/5$   & $100.0\%$ & $[56.6, 100.0]$ & $0.031$ \\
C          & $7/10$  & $70.0\%$  & $[39.7, 89.2]$ & $0.172$ \\
D          & $8/12$  & $66.7\%$  & $[39.1, 86.2]$ & $0.194$ \\
E          & $16/22$ & $72.7\%$  & $[51.8, 86.8]$ & $0.026$ \\
F          & $4/5$   & $80.0\%$  & $[37.6, 96.4]$ & $0.188$ \\
G          & $15/21$ & $71.4\%$  & $[50.0, 86.2]$ & $0.039$ \\
\midrule
\textbf{Pooled} & $\mathbf{67/93}$ & $\mathbf{72.0\%}$ & $\mathbf{[62.2, 80.1]}$ & $\mathbf{< 10^{-4}}$ \\
\bottomrule
\end{tabular}
\label{tab:pairwise-validation}
\end{table}

\subsection{Pointwise Validation via Likert Ratings: Acceptance-Anchored and
Acceptance-Independent Items}
\label{app:downstream-validation}
Whereas the pairwise study tests whether the metric \emph{separates} a chosen utterance from a rejected one, this study tests whether AUC \emph{monotonically tracks} human judgment of mediation effect across the full range: we sampled $30$ scenarios balanced across the three ethnocentric stages---$15$ from the high-AUC bucket (mean AUC $+2.81$) and $15$ from the low (mean AUC $-0.86$), all from CC-Mediation-SFT on the Llama-3.1-8B backbone---and asked $N=6$ raters, on a $5$-point Likert scale, whether they would come to view the other speaker's cultural perspective as \emph{``real, legitimate, and equally valid''} (the verbatim DMIS Acceptance definition) after hearing the mediator's utterance. The sample-level mean ratings correlate continuously with AUC at Spearman $\rho = +0.637$ (bootstrap $95\%$ CI $[+0.43, +0.75]$, $p = 1.6\!\times\!10^{-4}$; Table~\ref{tab:downstream-corr}), with signed $W_1$ showing a comparable $\rho = +0.562$, and leave-one-rater-out folds preserve the correlation ($\rho \in [+0.529, +0.686]$, $p < 0.01$ throughout). Together with the three preceding validation lines, this closes the loop on metric validity: the labeler is accurate at the turn level, the joint criterion separates better mediations from worse in pairwise comparison, and the metrics align continuously with human judgment of post-intervention Acceptance-shift likelihood.
\begin{table}[H]
\centering
\small
\caption{Sample-level correlation ($n=30$) between trajectory metrics and the human mean Likert rating, averaged across $6$ raters per sample. Confidence intervals from bootstrap with $10{,}000$ resamples.}
\setlength{\tabcolsep}{4pt}
\begin{tabular}{lcc}
\toprule
 & \textbf{Trajectory AUC} & \textbf{Signed $W_1$} \\
\midrule
Spearman $\rho$ & $\mathbf{+0.637}$ & $+0.562$ \\
$95\%$ CI       & $[+0.43, +0.75]$  & $[+0.28, +0.72]$ \\
Pearson $r$     & $+0.678$          & $+0.602$ \\
$p$             & $1.6\!\times\!10^{-4}$ & $1.2\!\times\!10^{-3}$ \\
\bottomrule
\end{tabular}
\label{tab:downstream-corr}
\end{table}

Because the anchored item quotes the DMIS Acceptance definition verbatim, its
correlation with AUC could in principle arise from measuring the same
definition twice. Instrument validation of a defined construct requires the
rater and the metric to target the same construct, so the anchored design is
intentional; to test convergence with a theory-independent outcome intuition,
however, we reran the study under the identical setup (same $30$ samples,
$N{=}6$ raters, $5$-point scale), replacing only the item with an
Acceptance-independent, outcome-oriented question: \emph{``Did this
intervention help the two speakers better understand each other's
positions?''} The correlation with AUC persists ($\rho = +0.372$,
$p = 0.043$; high- vs.\ low-AUC means $3.62$ vs.\ $3.31$), weaker than under
the anchored item ($+0.637$; $3.38$ vs.\ $2.36$) as expected, but
significant once the Acceptance vocabulary is entirely removed
(Table~\ref{tab:item-vocab}). The anchored correlation is therefore not
solely an artifact of shared wording: AUC also tracks an independent,
outcome-level notion of mediation effect, albeit more weakly than the
DMIS-defined construct it is built to measure.
\begin{table}[H]
\centering\small
\caption{AUC--human correlation by Likert-item vocabulary (identical samples,
raters, and scale). The anchored item is the original of
Table~\ref{tab:downstream-corr}; the independent item removes all DMIS
Acceptance vocabulary.}
\setlength{\tabcolsep}{4pt}
\begin{tabular}{lcc}
\toprule
 & Acceptance- & Acceptance- \\
 & anchored & independent \\
\midrule
$\rho$ vs.\ AUC & $+0.637$ & $+0.372$ \\
$p$             & $<0.001$ & $0.043$ \\
High-AUC mean   & $3.38$   & $3.62$ \\
Low-AUC mean    & $2.36$   & $3.31$ \\
$\Delta$        & $+1.02$  & $+0.31$ \\
\bottomrule
\end{tabular}
\label{tab:item-vocab}
\end{table}

\begin{table*}[t]
\centering
\caption{\textbf{Cross-Model Robustness of the chosen-side admission.} Seven of the eight $(M, C, L)$ cells of a $2{\times}2{\times}2$ matrix that independently varies the mediator $M$, the continuation simulator $C$, and the labeler $L$. The cross-family pairs (GPT-4o-mini $\leftrightarrow$ Claude-3.5-Haiku, GPT-4o-mini logprobs $\leftrightarrow$ Qwen3-14B local logits) test whether the chosen mediation's downstream effect survives when each role is replaced with a model from a different family. Trajectory AUC and signed $W_1$ are both strictly positive in every completed cell, and the Judge score---which uses a fixed GPT-4o-mini rubric judge independent of $L$---stays above $4.3$ throughout. AUC and $W_1$ use the uniformly spaced developmental scale ($[-5, +5]$); Judge is a $1$--$5$ LLM-judge rating.}
\small
\setlength{\tabcolsep}{6pt}
\begin{tabular}{l l l c c c}
\toprule
\textbf{Mediator $M$} & \textbf{Continuation $C$} & \textbf{Labeler $L$ (logit source)} & \textbf{AUC} & \textbf{$W_1$} & \textbf{Judge} \\
\midrule
GPT-4o-mini      & GPT-4o-mini      & GPT-4o-mini (API)            & $+0.82$ & $+0.83$ & $4.55$ \\
GPT-4o-mini      & GPT-4o-mini      & Qwen3-14B (local logits)     & $+0.72$ & $+0.96$ & $4.60$ \\
GPT-4o-mini      & Claude-3.5-Haiku & GPT-4o-mini (API)            & $+0.83$ & $+1.00$ & $4.59$ \\
GPT-4o-mini      & Claude-3.5-Haiku & Qwen3-14B (local logits)     & $+0.71$ & $+1.13$ & $4.64$ \\
Claude-3.5-Haiku & GPT-4o-mini      & GPT-4o-mini (API)            & $+0.90$ & $+0.96$ & $4.39$ \\
Claude-3.5-Haiku & GPT-4o-mini      & Qwen3-14B (local logits)     & $+0.75$ & $+1.05$ & $4.31$ \\
Claude-3.5-Haiku & Claude-3.5-Haiku & GPT-4o-mini (API)            & $+0.78$ & $+0.97$ & $4.38$ \\
\bottomrule
\end{tabular}
\label{tab:cross-model}
\end{table*}

\section{Additional Experiments}
\label{app:additional-experiments}

\begin{table*}[h]
\centering
\small
\caption{Final-layer values for the utterance-content probe.
Across three backbones the rubric vs.\ no-rubric gap is bounded by
$|\Delta^{\text{utt}}_{\text{elic}}|\!\le\!0.02$ on cosine and
$\le\!0.07$ on $R^2$, at both positions and of consistently
\emph{negative} sign---an order of magnitude below, and opposite in
sign to, the stage-channel
$\Delta_{\text{elic}}\!\approx\!+0.15$. The Utterance-end $>$
Prompt-end ordering on cosine (gap $0.08$--$0.13$) confirms the
probe is responsive to in-context utterance content.}
\setlength{\tabcolsep}{4.5pt}
\renewcommand{\arraystretch}{1.10}
\begin{tabular}{l l c c c c c c}
\toprule
 & & \multicolumn{3}{c}{\textbf{Cosine similarity}}
   & \multicolumn{3}{c}{\textbf{Coefficient of determination $R^2$}} \\
\cmidrule(lr){3-5}\cmidrule(lr){6-8}
\textbf{Backbone} & \textbf{Position}
  & \textbf{D} & \textbf{ND} & $\Delta^{\text{utt}}_{\text{elic}}$
  & \textbf{D} & \textbf{ND} & $\Delta^{\text{utt}}_{\text{elic}}$ \\
\midrule
\multirow{2}{*}{\textsc{Llama-3.1-8B-Instruct}}
  & Prompt-end     & 0.475 & 0.487 & $-0.012$ & $-0.431$ & $-0.382$ & $-0.049$ \\
  & Utterance-end  & 0.565 & 0.566 & $-0.001$ & $-0.161$ & $-0.144$ & $-0.018$ \\
\midrule
\multirow{2}{*}{\textsc{Gemma-2-9B-it}}
  & Prompt-end     & 0.527 & 0.538 & $-0.011$ & $-0.268$ & $-0.214$ & $-0.054$ \\
  & Utterance-end  & 0.602 & 0.622 & $-0.020$ & $-0.045$ & $-0.000$ & $-0.045$ \\
\midrule
\multirow{2}{*}{\textsc{Phi-3.5-mini-instruct}}
  & Prompt-end     & 0.467 & 0.483 & $-0.016$ & $-0.487$ & $-0.416$ & $-0.071$ \\
  & Utterance-end  & 0.607 & 0.616 & $-0.009$ & $-0.031$ & $-0.016$ & $-0.015$ \\
\midrule
\multicolumn{2}{r}{\emph{Stage-channel reference}}
  & \multicolumn{6}{l}{$\Delta_{\text{elic}}\!\approx\!+0.15$ at
    Utterance-end (Fig.~\ref{fig:layerwise})} \\
\bottomrule
\end{tabular}
\label{tab:utterance-probe}
\end{table*}

\subsection{Cross-Model Robustness}
\label{sec:cross-model}
A natural concern with a fully synthetic pipeline driven by a single LLM family (GPT-4o-mini) is that the reported mediation effect might reflect a self-reinforcing loop in which the same model family that generates the data also evaluates it. We address this by re-running the effect evaluation under a $2{\times}2{\times}2$ matrix that independently varies the mediator $M$, the continuation simulator $C$ (each between GPT-4o-mini and Claude-3.5-Haiku), and the labeler $L$ (between GPT-4o-mini API \texttt{logprobs} and \textbf{Qwen3-14B-Instruct} local logits). Across the seven completed cells in Table~\ref{tab:cross-model}, swapping the labeler to Qwen3-14B lowers AUC by $0.10$--$0.15$ but leaves every cell strictly positive ($+0.71$ to $+0.75$) while $W_1$ actually rises ($+0.96$ to $+1.13$); swapping the simulator and mediator to Claude-3.5-Haiku preserves both the direction and magnitude of post-intervention trajectories with Judge above $4.3$ throughout, ruling out three distinct self-reinforcement hypotheses against a single-model synthetic pipeline.

\subsection{Utterance-Content Probe}
\label{app:utterance-probe}
To rule out a ``rubric preserves next-utterance content'' reading of Fig.~\ref{fig:layerwise}, we re-use the hidden states from the main stage probe ($N{=}158$ scenarios; $2{\times}2$ over position $\in$\{Prompt-end, Utterance-end\} crossed with condition $\in$\{D, ND\}; three open-weight backbones) and swap the target from the 3-class DMIS stage to the 384-dim sentence embedding of the gold mediator utterance (\texttt{all-MiniLM-L6-v2}), fitting per-(model, position, condition, layer) ridge regression under the same scenario-level 5-fold CV (Table~\ref{tab:utterance-probe}). The probe is responsive---Utterance-end cosine exceeds Prompt-end by $0.08$--$0.13$---but the condition effect is null: $\Delta^{\text{utt}}_{\text{elic}}\!\equiv\!\text{cos}(D)\!-\!\text{cos}(ND)$ is bounded by $|\Delta|\!\le\!0.02$ at all six (model, position) cells, always \emph{negative}, and an order of magnitude below the stage-channel $\Delta_{\text{elic}}\!\approx\!+0.15$ at Utterance-end (Fig.~\ref{fig:layerwise}), supporting the differential claim that there is no rubric-dependent utterance-content gap and thereby disambiguating rubric-as-stage-anchor from rubric-as-content-anchor in favor of the former.

\begin{figure*}[ht]
    \centering
    \includegraphics[width=\textwidth]{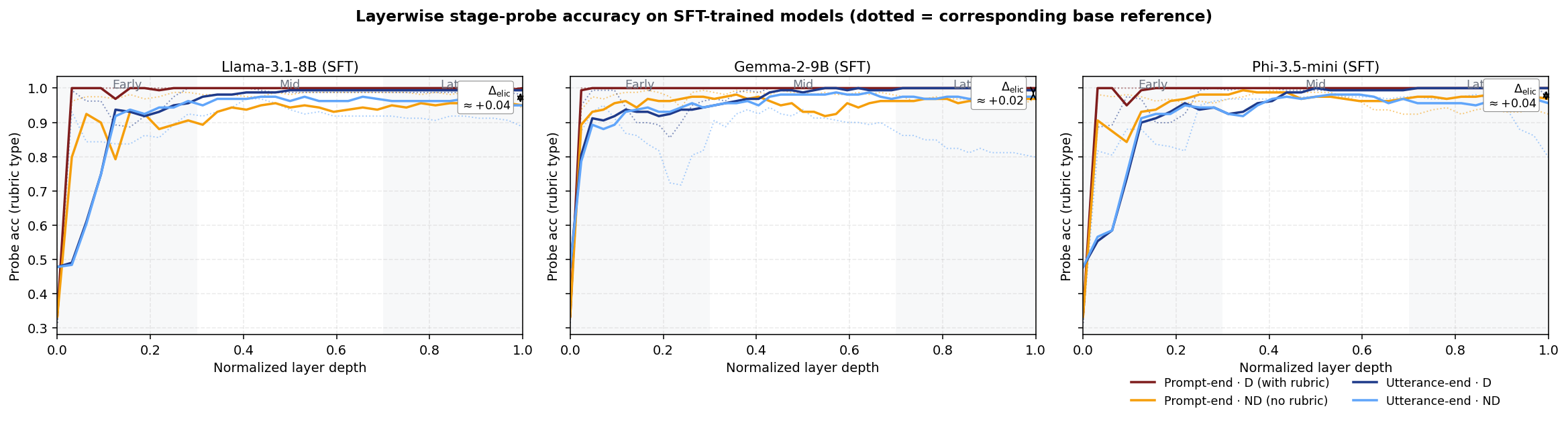}
    \caption{Layerwise DMIS stage decoding accuracy on SFT-trained models (Llama-3.1-8B, Gemma-2-9B, Phi-3.5-mini)}
    \label{fig:layerwise-sft-vs-base}
\end{figure*}

\begin{figure}[ht]
    \centering
    \includegraphics[width=\linewidth]{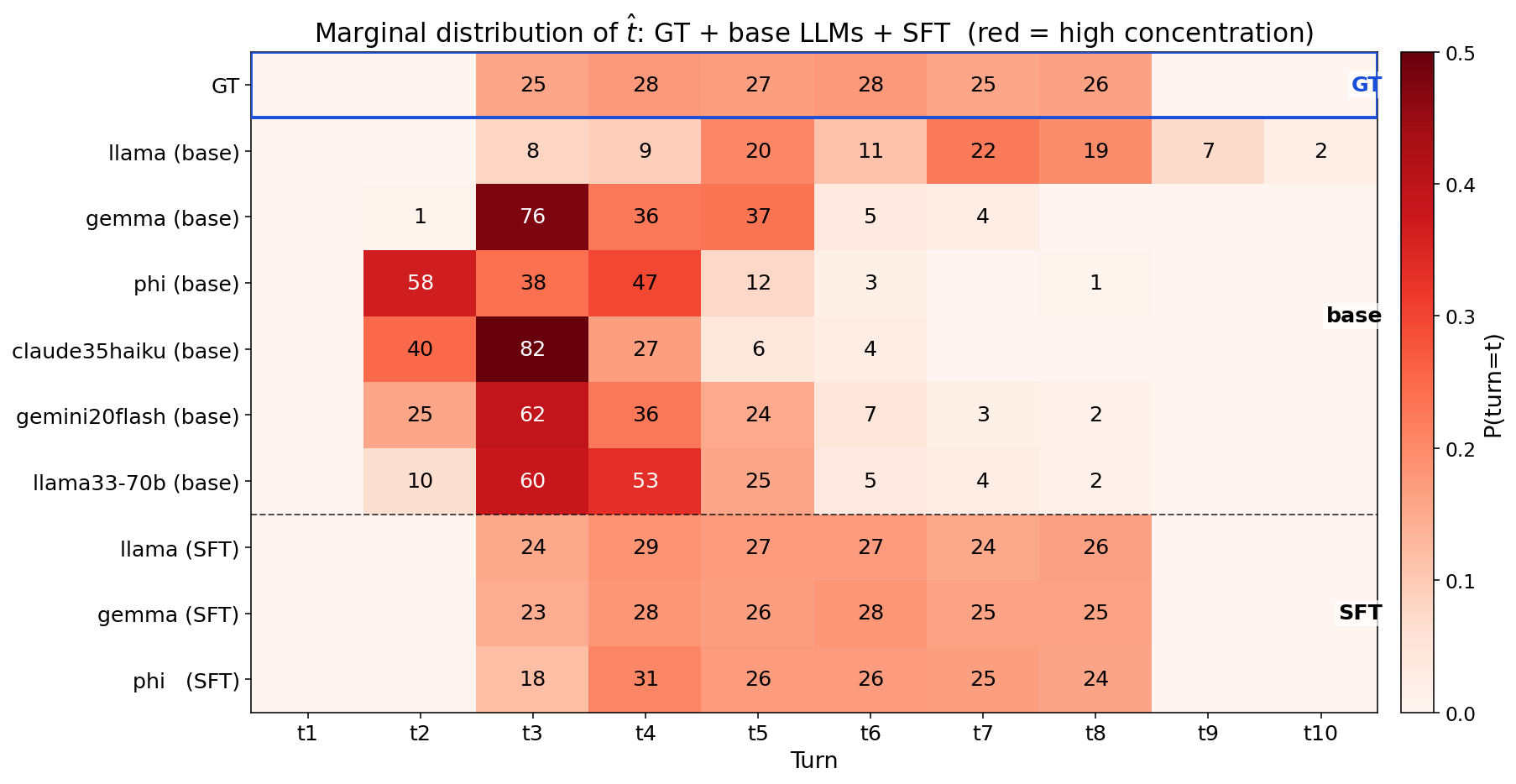}
    \caption{Marginal distribution of $\hat{t}$ for ground truth (GT), base LLMs, and SFT-trained models}
    \label{fig:h2-marginal-distribution}
\end{figure}

\subsection{Effectiveness of FT Model}
\paragraph{Does SFT Mitigate Positional Bias?}
\label{subsec:sft-positional-bias}

Figure~\ref{fig:h2-marginal-distribution} compares the marginal distribution of $\hat{t}$ across ground truth (GT), base LLMs, and SFT-trained models, directly revealing the \textit{positional bias} that base LLMs commonly exhibit. The GT distribution is nearly uniform across $t_3$ through $t_8$, with each turn receiving between $25\%$ and $28\%$ of the probability mass. In contrast, all base models concentrate their predictions on a few early positions in the sequence. The concentration at $t_3$ is particularly striking: \texttt{claude-3.5-haiku} assigns $82\%$, \texttt{gemma-2-9B} $76\%$, \texttt{gemini-2.0-flash} $62\%$, and \texttt{llama-3.3-70B} $60\%$, while later turns ($t_6$ onward) receive almost no probability mass. This bias is observed consistently across model scales (8B--70B) and families (Llama, Gemma, Phi, Claude, Gemini), indicating a systematic positional bias rather than a model-specific artifact.

After SFT, the marginal distributions of all three trained models (\texttt{llama}, \texttt{gemma}, \texttt{phi}) realign almost perfectly with the GT. Their probability values over $t_3$--$t_8$ fall within $24$--$29\%$, $23$--$28\%$, and $18$--$31\%$ respectively, closely matching the GT range of $25$--$28\%$. The probability mass that was previously concentrated on early turns is redistributed evenly across later turns. These results show that SFT effectively removes the early-turn positional bias shared by base LLMs and restores the positional calibration of the predicted distribution.

\paragraph{Does SFT Resolve Late-Layer Representation Decay?}
\label{subsec:sft-effect}

Figure~\ref{fig:layerwise-sft-vs-base} compares the layerwise stage-probe accuracy of base and SFT-trained models under four conditions (Prompt-end$\cdot$D, Prompt-end$\cdot$ND, Utterance-end$\cdot$D, Utterance-end$\cdot$ND), illustrating the \textit{late-layer representation decay} observed in base models and its recovery through SFT. For the base models (dotted lines), stage-relevant information is encoded reasonably well in the early and middle layers, but probe accuracy declines noticeably as the normalized layer depth exceeds $0.6$. This decay is especially severe under the ND (no rubric) condition: the Utterance-end$\cdot$ND curve (light blue dotted) of \texttt{Gemma-2-9B} drops to roughly $0.80$ in the deepest layers, and \texttt{Phi-3.5-mini} exhibits a similar late-layer degradation. This pattern suggests that task-relevant stage information is gradually lost as it propagates into deeper layers.

After SFT (solid lines), the late-layer representation decay is effectively eliminated across all three models. Probe accuracy under all four conditions reaches a plateau above $0.95$ from the middle layers onward, and the downward slope in the deeper layers observed in the base models flattens out. In particular, the most fragile Utterance-end$\cdot$ND condition recovers to a level comparable to its D counterpart, yielding consistent improvements in the elicitation gap: $\Delta_{\text{elic}} \approx +0.04$ for \texttt{Llama-3.1-8B}, $+0.02$ for \texttt{Gemma-2-9B}, and $+0.04$ for \texttt{Phi-3.5-mini}. Overall, SFT resolves the representation decay observed in the deeper layers of base models, ensuring that task-relevant information is stably preserved and expressed throughout the network.

\subsection{Generalization under Distribution Shift}
\label{app:dist-shift}
The SFT timing gain ($32.7\% \rightarrow 98.1\%$ on Llama-3.1-8B) is large on
a corpus built from a single scenario source, raising the concern that the
model fits the SocialCC/GPT-4o-mini template rather than learning
conflict-reading. We test this on all three shift axes: we re-partition the
full $1{,}661$-scenario pool, entirely remove the held-out scenarios from
training (no partial leakage), and retrain all three backbones ($9$
configurations): \textbf{(a)}~an unseen-country split withholding every
scenario involving China, Canada, or Japan ($n{=}905$); \textbf{(b)}~an
unseen-WVS-dimension split withholding the entire Q19 (neighbor-tolerance)
dimension ($n{=}330$); and \textbf{(c)}~an unseen-template split withholding
the five most frequent scenario texts ($n{=}222$).

Table~\ref{tab:dist-shift} shows that on the two larger backbones, the
WVS-dimension and template holdouts preserve turn-acc at essentially the
in-distribution level ($95.2$--$97.9\%$ vs.\ $98.1\%$): the model fits
neither a specific value dimension nor the surface textual form of the
scenarios. The country holdout---the one axis whose removal changes cultural
content rather than surface form---is the only axis that degrades: timing
holds at $79$--$84\%$ while Judge drops to $2.0$--$2.1$. This localization is
itself evidence against template-fitting: had the model memorized surface
templates, performance would persist as long as the template is unchanged,
regardless of which countries appear. The model instead depends on
country-level cultural context, and the quality drop reflects reduced
coverage of the withheld cultures---addressable by broadening cultural
coverage rather than a sign of surface overfitting. Phi-3.5-mini, already the
weakest backbone in-distribution, degrades further under every shift,
consistent with its capacity limitation rather than as a separate anomaly.
\begin{table}[H]
\centering\footnotesize
\caption{Held-out distribution-shift results. Held-out scenarios ((a)~Country
$n{=}905$, (b)~WVS-dim $n{=}330$, (c)~Template $n{=}222$, identical across
backbones) are entirely removed from training and each backbone is retrained
per axis ($9$ configurations). Reference: in-distribution SFT turn-acc
$= 98.1\%$ (Llama-3.1-8B).}
\setlength{\tabcolsep}{3pt}
\resizebox{\columnwidth}{!}{%
\begin{tabular}{llcccc}
\toprule
Backbone & Holdout & TurnAcc & AUC & $W_1$ & Judge \\
\midrule
\multirow{3}{*}{Llama-3.1-8B}
 & (b) WVS-dim   & $\mathbf{97.9}$ & $1.00$ & $0.95$ & $3.6$ \\
 & (a) Country   & $84.2$          & $0.70$ & $0.76$ & $2.0$ \\
 & (c) Template  & $\mathbf{96.8}$ & $0.74$ & $0.83$ & $3.8$ \\
\midrule
\multirow{3}{*}{Gemma-2-9B}
 & (b) WVS-dim   & $\mathbf{95.2}$ & $0.93$ & $0.88$ & $4.1$ \\
 & (a) Country   & $79.4$          & $0.70$ & $0.86$ & $2.1$ \\
 & (c) Template  & $\mathbf{97.3}$ & $0.79$ & $0.80$ & $4.1$ \\
\midrule
\multirow{3}{*}{Phi-3.5-mini}
 & (b) WVS-dim   & $76.1$ & $0.77$ & $0.83$ & $1.8$ \\
 & (a) Country   & $39.0$ & $0.66$ & $0.60$ & $1.6$ \\
 & (c) Template  & $73.9$ & $0.78$ & $0.67$ & $2.2$ \\
\bottomrule
\end{tabular}%
}
\label{tab:dist-shift}
\end{table}

\subsection{Timing-Matched Paired Comparison of Base and SFT}
\label{app:timing-matched}
Because $\text{AUC}_{\text{exact}}$ averages over very different sample sizes
before and after SFT (\S\ref{sec:rq3}), we re-compare Base and SFT on the
identical subset of scenarios in which \emph{both} intervened exactly at
$t^{*}$, using a paired two-sided Wilcoxon signed-rank test
(Table~\ref{tab:timing-matched}). On identical scenarios the apparent
degradation disappears: every AUC delta is positive, and no negative delta is
distinguishable from zero (Phi $W_1$: $\Delta{-}0.09$, $p{=}1.00$). The
trajectory deltas are directionally positive but underpowered at
$n{=}17$--$52$, and we do not claim significance for them. Judge---the metric
least exposed to the timing confound, since it scores the single utterance
against the stage rubric rather than a downstream trajectory---rises strongly
and significantly on the two larger backbones ($+1.79$ and $+1.15$).
Phi-3.5-mini shows no detectable content change in either direction.
\begin{table}[H]
\centering\footnotesize
\caption{Base vs.\ SFT on the identical scenarios where both intervened at
$t^{*}$ (paired, two-sided Wilcoxon signed-rank). Bold: significant at
$p<0.001$.}
\setlength{\tabcolsep}{3pt}
\resizebox{\columnwidth}{!}{%
\begin{tabular}{llccc}
\toprule
 & & Llama-3.1-8B & Gemma-2-9B & Phi-3.5-mini \\
 & & ($n{=}52$) & ($n{=}34$) & ($n{=}17$) \\
\midrule
\multirow{2}{*}{AUC}
 & Base$\rightarrow$SFT & $+0.52 \rightarrow +0.67$ & $+0.68 \rightarrow +0.86$ & $+0.43 \rightarrow +0.49$ \\
 & $\Delta$ ($p$) & $+0.16$ ($.30$) & $+0.18$ ($.24$) & $+0.06$ ($.67$) \\
\midrule
\multirow{2}{*}{$W_1$}
 & Base$\rightarrow$SFT & $+0.65 \rightarrow +0.77$ & $+0.71 \rightarrow +0.75$ & $+0.43 \rightarrow +0.34$ \\
 & $\Delta$ ($p$) & $+0.12$ ($.15$) & $+0.05$ ($.44$) & $-0.09$ ($1.00$) \\
\midrule
\multirow{2}{*}{Judge}
 & Base$\rightarrow$SFT & $2.90 \rightarrow 4.69$ & $3.06 \rightarrow 4.21$ & $2.47 \rightarrow 2.76$ \\
 & $\Delta$ ($p$) & $\mathbf{+1.79}$ ($1.5{\times}10^{-8}$) & $\mathbf{+1.15}$ ($4.6{\times}10^{-5}$) & $+0.29$ ($.55$) \\
\bottomrule
\end{tabular}%
}
\label{tab:timing-matched}
\end{table}

\subsection{Evaluation-GT Performance by Filter Outcome}
\label{app:gt-filter-split}
The Step-4 filter is applied only to training mediations
(\S\ref{sec:benchmark}); applying it to the evaluation set would remove the
lower-quality tail and inflate the reference. For completeness, we report the
effect of applying the same gate condition (the supervised mediation strictly
outperforms its unguided baseline on both AUC and $W_1$) to the $158$
evaluation GT mediations: $41$ pass ($25.9\%$).
Table~\ref{tab:gt-filter-split} splits GT performance by filter outcome. On
AUC and $W_1$ the passed subset is clearly separated from the failed subset
($+2.38$ vs.\ $+1.38$; $+1.80$ vs.\ $+1.45$), confirming that the unfiltered
mean of $1.64$ is a mixed-quality value pulled down by filter-failing cases,
not an attainable ceiling---hence the ``GT reference'' label in
Table~\ref{tab:rq2}. Judge is essentially flat across the split ($4.93$ vs.\
$4.95$): consistent with its definition, the gate conditions on the sign of
the AUC/$W_1$ improvement and is not a general quality selector, so the
passed subset serves specifically as an AUC/$W_1$ reference rather than a
global quality ceiling. Measured against this reference, the gap between
evaluated models (AUC $0.68$--$1.54$; Table~\ref{tab:rq2}) and GT-level
mediation is wider still.
\begin{table}[H]
\centering\small
\caption{Evaluation-GT performance split by the Step-4 gate condition,
applied post hoc for analysis only.}
\begin{tabular}{lcccc}
\toprule
Subset & $n$ & AUC & $W_1$ & Judge \\
\midrule
Filter-passed & $41$ ($25.9\%$) & $+2.38$ & $+1.80$ & $4.93$ \\
Filter-failed & $117$ ($74.1\%$) & $+1.38$ & $+1.45$ & $4.95$ \\
All (unfiltered) & $158$ & $+1.64$ & $+1.54$ & $4.94$ \\
\bottomrule
\end{tabular}
\label{tab:gt-filter-split}
\end{table}

\subsection{Marker-Level Analysis of the Gemma-2-9B Rubric Anomaly}
\label{app:gemma-marker}
On Gemma-2-9B, providing the rubric (ND$\rightarrow$D) moves AUC only
$0.72 \rightarrow 0.75$ while $W_1$ ($0.81 \rightarrow 0.78$) and Judge
($3.34 \rightarrow 3.20$) decline; per-scenario paired tests (Wilcoxon
signed-rank, two-sided, $n{=}158$) find none of the three deltas
distinguishable from zero (AUC $\Delta{+}0.047$, $p{=}.598$; $W_1$
$\Delta{-}0.024$, $p{=}.534$; Judge $\Delta{-}0.134$, $p{=}.206$), so the
apparent declines lie within paired noise and the rubric's measurable output
effect is negligible on this backbone.

To locate where the rubric's effect is lost, we derive per-stage marker
lexicons from the \texttt{CONFLICT\_TABLE} definitions and score each
generated mediator utterance for (i)~whether it contains markers of a
stage-specific mediation move and (ii)~whether those markers match the target
stage. Table~\ref{tab:gemma-marker} reports the ND$\rightarrow$D change in
marker-application rate for all six models. Five models move within
$[-5.7, +21.4]$\,pp, while Gemma-2-9B alone drops by $-20.8$\,pp
($82.4\% \rightarrow 61.6\%$); among its utterances that do carry markers
under D, $60.2\%$ ($59/98$) select a stage mismatched to the target. Given
that its internal stage representation is fully restored by the rubric
(utterance-end D probe $=1.00$; \S\ref{sec:rq2}), the anomaly is a failure of
surface realization during generation, not of representation---the
representation--output gap of \S\ref{sec:rq2} at its widest.
\begin{table}[H]
\centering\small
\caption{Change in stage-marker application rate when the rubric is provided
(ND$\rightarrow$D), by model.}
\begin{tabular}{lccr}
\toprule
Model & ND & D & $\Delta$ \\
\midrule
Claude-3.5-Haiku & $88.5\%$ & $99.4\%$ & $+10.9$\,pp \\
Llama-3.3-70B    & $92.4\%$ & $100\%$  & $+7.6$\,pp \\
Gemini-2.0-Flash & $92.4\%$ & $90.4\%$ & $-2.0$\,pp \\
Llama-3.1-8B     & $93.1\%$ & $87.4\%$ & $-5.7$\,pp \\
Phi-3.5-mini     & $67.9\%$ & $89.3\%$ & $+21.4$\,pp \\
\textbf{Gemma-2-9B} & $\mathbf{82.4\%}$ & $\mathbf{61.6\%}$ & $\mathbf{-20.8}$\,pp \\
\bottomrule
\end{tabular}
\label{tab:gemma-marker}
\end{table}


\section{Case Studies}
\label{app:case-studies}

We present one worked example per ethnocentric stage. Each case shows the full ten-turn dialogue: a shared prefix up to the conflict turn $t_{\text{conf}}$, the original continuation without mediation, and the mediated continuation produced by injecting the supervised mediator utterance at $t_{\text{conf}}$. All three cases are drawn from the \textsc{CC-Mediation} chosen path ($\texttt{turn\_gt\_w\_df}$) with judge score $5$, and the mediator across all three is \texttt{claude-3.5-haiku}. These are not cherry-picked outputs of the trained timing model; they are illustrations of the supervised target the model is trained against.

\subsection{Denial --- sid 388 ($t_{\text{conf}} = 8$)}
\label{app:case-denial}

\paragraph{Agents and setting.}
Kelly is a 17-year-old high-school student from the USA (\textsc{Other}); Omar is an 18-year-old high-school student from Mongolia (\textsc{Conflict}). They are picking a gift for their friend Adam's scholarship celebration. The cultural axis is individualist self-fulfilment (Kelly: ``I want to make myself happy and do my own thing'') vs.\ collectivist family-pride (Omar: ``making my parents proud is everything'').

\paragraph{Shared prefix ($t_1$--$t_8$).}
\begin{quote}\small
\noindent\textbf{$t_1$ [Kelly, \textsc{Other}]}: ``Omar, I was thinking, for Adam's celebration, we should get him a nice bottle of champagne. What do you think?''

\noindent\textbf{$t_2$ [Omar, \textsc{Conflict}]}: ``I think a traditional Mongolian gift would be more meaningful, like a nice cashmere scarf or a horse statue, something that represents our culture and shows respect to his achievement, my parents would be proud if I bring a thoughtful gift like that.''

\noindent\textbf{$t_3$ [Kelly]}: ``I'm not sure about that, Omar. Champagne is classic for celebrations and Adam would probably appreciate it. Besides, I don't really think about what my parents would think when it comes to my goals or actions, that's not really a driving force for me.''

\noindent\textbf{$t_4$ [Omar]}: ``For me, making my parents proud is everything, it's what drives me to succeed, and I think a traditional gift would show Adam that I put thought into it, my parents would be happy to know I'm respecting our traditions.''

\noindent\textbf{$t_5$ [Kelly]}: ``I don't know\ldots I just don't think making my parents proud is that big of a deal to me, I mean, I want to make myself happy and do my own thing. And I still think champagne is a better idea for Adam's party.''

\noindent\textbf{$t_6$ [Omar]}: ``I understand that you think differently, but for me, my parents' pride is everything, it's a big deal, and I will always consider their feelings and expectations, I think I'll go with the cashmere scarf.''

\noindent\textbf{$t_7$ [Kelly]}: ``I'm just not on the same page as you when it comes to that\ldots I don't think my goals in life have been to make my parents proud. Anyway, back to Adam's gift, are you sure he'd even like a scarf?''

\noindent\textbf{$t_8$ [Omar --- \denial{Denial}, $\star$]}: ``I don't really see meaningful differences here, people figure it out, and at the end of the day it's all the same, doesn't matter to me, so let's just live and let live, I think we should focus on getting a nice gift for Adam, as long as we're speaking the same language, we can make this work, and I'm sure with experience you can handle any situation.''
\end{quote}

\paragraph{Without mediation.}
\begin{quote}\small
\noindent\textbf{$t_9$ [Kelly --- \minimization{Minimization}]}: ``I still think champagne would be better, it's just more of a celebration thing, you know? And I don't think we should just settle on something because it's easy, I really think Adam would appreciate the champagne more. Can't we just get that instead?''

\noindent\textbf{$t_{10}$ [Omar --- \defense{Defense}]}: ``I'm not trying to settle, I just think the scarf is a more thoughtful gift, and it's what my parents would want me to get, they would be proud if I bring a traditional gift, and that's important to me, I'm not going to change my mind about this, a scarf is what I'm getting.''
\end{quote}

\paragraph{With mediation.}
\begin{quote}\small
\noindent\textbf{[Mediator $\rightarrow$ both]}: \emph{``I'm curious about something --- it seems like you both care deeply about showing respect to Adam, just in different ways. Omar, could you tell me a bit more about what the cashmere scarf symbolizes in your cultural tradition? And Kelly, what makes champagne feel like the right celebration gesture to you? I'd love to hear the stories behind your gift ideas.''}

\noindent\textbf{$t_9$ [Kelly]}: ``I think champagne symbolizes celebration and achievement. It's like a universal way to toast to someone's success, and I feel it matches the festive vibe of the party. To me, it's about celebrating Adam's hard work in a way that feels exciting and joyful.''

\noindent\textbf{$t_{10}$ [Omar --- \acceptance{Acceptance}]}: ``For me, the cashmere scarf symbolizes warmth and respect. In my culture, gifting something that represents our heritage shows appreciation for someone's achievements in a deeper way. It's a reminder of how important family and traditions are to me, and it would mean a lot to my parents to see me honoring those values while celebrating Adam.''
\end{quote}

\paragraph{Discussion.}
Turn~8 surfaces all four canonical Denial markers in a single utterance: ``I don't really see meaningful differences'' (denial of the cultural dimension), ``at the end of the day it's all the same'' (universalising flattening), ``doesn't matter to me'' (disengagement), and ``live and let live'' (the verbatim Denial closer). \emph{Without mediation}, the topic-levelling move does not stabilise the disagreement: Omar slides into outright Defense at $t_{10}$ (``I'm not going to change my mind'', ``a scarf is what I'm getting''), while Kelly drifts into Minimization at $t_9$ --- the conversation worsens along both axes. \emph{With mediation}, rather than addressing the Denial markers head-on, the mediator bypasses the topic-levelling move and re-routes both speakers onto the cultural-symbolism axis by asking each to \emph{narrate} what their preferred gift means. The result is the textbook Acceptance form at $t_{10}$: Omar holds the same preference (cashmere scarf) but now articulates it from \emph{inside} his cultural frame (``In my culture, gifting something that represents our heritage\ldots'') rather than dropping the cultural dimension. Same speaker, same preference, a different stance toward the cultural dimension of the disagreement.

\subsection{Defense --- sid 62 ($t_{\text{conf}} = 6$)}
\label{app:case-defense}

\paragraph{Agents and setting.}
Ryan is a 29-year-old geologist from Cyprus (\textsc{Other}); Mia is a 26-year-old architect from Singapore (\textsc{Conflict}). They are choosing a housewarming gift for their friend Laura and the conversation drifts to Laura's new neighbors. The cultural axis is Mediterranean party-friendly sociability (Ryan) vs.\ Singaporean preference for quiet, considerate neighbors (Mia). This is also the scenario rendered in Figure~\ref{fig:example}.

\paragraph{Shared prefix ($t_1$--$t_6$).}
\begin{quote}\small
\noindent\textbf{$t_1$ [Ryan, \textsc{Other}]}: ``I think white lilies would be a great housewarming gift for Laura, don't you, Mia? They're elegant and symbolize purity and innocence, which would be a nice touch for her new apartment.''

\noindent\textbf{$t_2$ [Mia, \textsc{Conflict}]}: ``White lilies are a good choice, but I was thinking perhaps something more practical, like a kitchen appliance or a personalized item for her new space. That way, she can really use it and remember us by it.''

\noindent\textbf{$t_3$ [Ryan]}: ``Practical gifts are thoughtful, but I still think white lilies would be a nice gesture. By the way, have you met any of Laura's new neighbors yet, and do you think it's a good thing that some of them like to drink heavily?''

\noindent\textbf{$t_4$ [Mia]}: ``Honestly, I'm not comfortable with the idea of heavy drinkers as neighbors. It can get noisy and disruptive. I think it's better to have considerate and respectful people living nearby. In Singapore, we value a more subdued and polite living environment.''

\noindent\textbf{$t_5$ [Ryan]}: ``I understand where you're coming from, Mia, but I think having heavy drinkers as neighbors can actually be a good thing. In Cyprus, we enjoy socializing and having a good time, and heavy drinkers can definitely liven up the atmosphere. It's all about being friendly and sociable, right?''

\noindent\textbf{$t_6$ [Mia --- \defense{Defense}, $\star$]}: ``No, that's not how it works. In our culture, we value respect for others and their space, and heavy drinking is NOT considered a desirable trait in neighbors. Actually, it's seen as immature and irresponsible. Properly speaking, a good neighbor is someone who is considerate and quiet, NOT someone who disrupts the peace with loud parties and drunken behavior.''
\end{quote}

\paragraph{Without mediation.}
\begin{quote}\small
\noindent\textbf{$t_7$ [Ryan --- Adaptation]}: ``I see things differently, Mia. In Cyprus, we \emph{embrace} a lively atmosphere and having heavy drinkers as neighbors can be a plus, it's all about being open-minded and welcoming to different personalities, and I think Laura's new neighbors will bring a fun dynamic to the building.''

\noindent\textbf{$t_8$ [Mia --- \defense{Defense}]}: ``I strongly disagree. Having heavy drinkers as neighbors is not something to be taken lightly. It's not about being open-minded, it's about having a decent living environment. I would never want to live next to someone who drinks heavily, it's just not acceptable to me.''

\noindent\textbf{$t_9$ [Ryan --- \minimization{Minimization}]}: ``I still think it's a non-issue, Mia. People in Cyprus enjoy their drinks and know how to have a good time without disturbing others, and I believe Laura's new neighbors will be fine, they might even invite us over for a barbecue or something.''

\noindent\textbf{$t_{10}$ [Mia --- \defense{Defense}]}: ``I don't think that's something to look forward to. Heavy drinking is not something to be encouraged or participated in, especially in a residential area. It's not about having a good time, it's about being respectful of those around you. I would not be comfortable attending a gathering like that, and I think Laura should be careful about getting too close to neighbors like that. GOOD BYE!''
\end{quote}

\paragraph{With mediation.}
\begin{quote}\small
\noindent\textbf{[Mediator $\rightarrow$ both]}: \emph{``I hear that you both care deeply about creating a positive living environment and want the best for Laura. Could we explore what qualities make a good neighbor from both of your perspectives? What shared values might you both see as important when it comes to community and mutual respect?''}

\noindent\textbf{$t_7$ [Ryan --- \acceptance{Acceptance}]}: ``I appreciate the mediator stepping in. I still believe that having neighbors who enjoy drinking can bring a sense of community and fun. In Cyprus, social gatherings are a big part of our culture, and it often leads to forming strong bonds. That said, I think we can both agree that a friendly atmosphere is important for Laura's new home, no matter our different views on drinking.''

\noindent\textbf{$t_8$ [Mia --- \acceptance{Acceptance}]}: ``I agree that a friendly atmosphere is important, but I still believe it should come from neighbors who are respectful and responsible. In Singapore, we focus on building a community where everyone feels comfortable and safe, and that often means avoiding excess drinking. It's great to have fun, but it shouldn't come at the cost of peace and respect for others.''

\noindent\textbf{$t_9$ [Ryan --- \acceptance{Acceptance}]}: ``I see your point about wanting a respectful environment, but I think there's room for both fun and respect. In Cyprus, we find ways to enjoy life while still being considerate. It's all about balance, wouldn't you say? Maybe we can find common ground on what makes a good neighbor and still celebrate our differences in perspective.''

\noindent\textbf{$t_{10}$ [Mia --- \acceptance{Acceptance}]}: ``Balance is important, but I still maintain that heavy drinking often disrupts that balance. A good neighbor should prioritize the comfort of others, and excessive drinking tends to clash with that. It's crucial to have neighbors who can enjoy themselves without compromising the peace. We can celebrate differences, but a respectful living environment is essential.''
\end{quote}

\paragraph{Discussion.}
Turn~6 stacks all four canonical Defense markers into a single utterance: flat contradiction (``No, that's not how it works''), us-vs-them (``In our culture''), evaluative all-caps (``NOT''), corrective opener (``Actually\ldots''), and prescriptive normative (``Properly speaking\ldots''). \emph{Without mediation}, Defense proves harder to release than Denial: Ryan's two attempts at conciliation ($t_7$ ``I see things differently'' and $t_9$ ``it's a non-issue'') are both met with sustained Defense from Mia ($t_8$, $t_{10}$), and the conversation terminates with ``GOOD BYE!'' --- any move that re-emphasises cultural contrast reinforces rather than de-escalates the polarisation. \emph{With mediation}, the mediator moves the conversation off the binary it had been stuck on (``heavy drinkers are good neighbors'' vs.\ ``\ldots are bad neighbors'') and onto a third axis (``what \emph{qualities} make a good neighbor''), naming the shared concern (``you both care deeply about creating a positive living environment'') before posing the question --- absorbing the conflict's emotional charge without recapitulating the cultural contrast itself. Mia's $t_{10}$ holds the same substantive position as her $t_6$ Defense (heavy drinking is incompatible with good neighbouring), but the form is no longer evaluative-prescriptive (``Properly speaking\ldots NOT'') and instead articulates the \emph{why} (``A good neighbor should prioritize the comfort of others''); the argmax shifts to Acceptance for all four post-intervention turns.

\subsection{Minimization --- sid 359 ($t_{\text{conf}} = 4$)}
\label{app:case-minimization}

\paragraph{Agents and setting.}
Adam is a 20-year-old Chinese law-school student (\textsc{Other}); Ali is a 19-year-old Canadian med-school student (\textsc{Conflict}). They are picking a gift for their friend Sarah who has just been accepted to medical school. The cultural axis is Canadian family-pride (Ali: ``making our parents proud\ldots that's just a universal value'') vs.\ a more individual-centric Chinese frame (Adam: ``I don't necessarily believe that making my parents proud is my main goal''). Note that the conflict agent here universalises \emph{toward} the family-pride frame, treating his own cultural value as the neutral baseline.

\paragraph{Shared prefix ($t_1$--$t_4$).}
\begin{quote}\small
\noindent\textbf{$t_1$ [Adam, \textsc{Other}]}: ``Hey Ali, have you thought about what gift we should get for Sarah's celebration? I was thinking maybe some frangipanis, they're a nice gesture, right?''

\noindent\textbf{$t_2$ [Ali, \textsc{Conflict}]}: ``Uh, flowers are okay, I guess\ldots but I was thinking something more personal, like a medical textbook or a customized stethoscope. That would be more meaningful for a future doctor, don't you think?''

\noindent\textbf{$t_3$ [Adam]}: ``But frangipanis symbolize good luck and new beginnings, which is perfect for her new journey in medical school. Plus, they're not as\ldots practical as a textbook, but more of a heartfelt gesture, you know?''

\noindent\textbf{$t_4$ [Ali --- \minimization{Minimization}, $\star$]}: ``I think we're getting caught up in the details, but deep down, we are all the same, and people are pretty much motivated by the same things. When it comes to making our parents proud, I mean, that's just a universal value, right? We both know, at the end of the day, that's what drives us. Some values are just universal, and I think this is one of them. It's a small world after all, and underneath we are the same. So, let's focus on finding a gift that comes from the heart, like you said, and I'm sure Sarah will appreciate it.''
\end{quote}

\paragraph{Without mediation.}
\begin{quote}\small
\noindent\textbf{$t_5$ [Adam --- \acceptance{Acceptance}]}: ``I don't think that's true for me, though\ldots I mean, making my parents proud isn't really a driving force for me. I'm not sure I agree that it's a universal value, either. Can we focus on the gift idea instead? Maybe the frangipanis could be a nice addition to a larger gift, like a celebration basket or something?''

\noindent\textbf{$t_6$ [Ali --- \defense{Defense}]}: ``I still think a practical gift is the way to go, but if you want to add some flowers to it, that's fine. For me, making my parents proud is \emph{everything}, and I think it should be for everyone. It's just the right thing to do, you know?''

\noindent\textbf{$t_7$ [Adam --- \acceptance{Acceptance}]}: ``I understand where you're coming from, but for me, it's just not that simple. Making my parents proud isn't\ldots it's not what motivates me, okay? Let's just drop it. So, are you on board with the celebration basket idea, or do you have a better idea for a practical gift?''

\noindent\textbf{$t_8$ [Ali --- \defense{Defense}]}: ``I'm on board with the basket, I guess. But just to clarify, I'm not going to change my mind about what's important --- making my parents proud is a top priority for me, and I think it should be for everyone.''

\noindent\textbf{$t_9$ [Adam --- \acceptance{Acceptance}]}: ``Those are good ideas for the basket. Let's just focus on making it a nice celebration for Sarah, okay? I'm not going to change my mind about what drives me, either\ldots and it's not about making my parents proud. Can we just agree to disagree on that and move forward with the party planning?''

\noindent\textbf{$t_{10}$ [Ali --- \defense{Defense}]}: ``Yeah, let's just focus on the party then. I think we've got a good idea with the basket, and like I said, a medical dictionary or notecards would be a nice touch. My parents would be thrilled to see me giving a thoughtful gift like that, and that means a lot to me.''
\end{quote}

\paragraph{With mediation.}
\begin{quote}\small
\noindent\textbf{[Mediator $\rightarrow$ both]}: \emph{``I'm hearing that you both want to honor Sarah's achievement, but I'm curious --- could you each share a bit about how gift-giving and celebration are typically approached in your own cultural backgrounds? What would your respective families consider meaningful in this kind of milestone moment?''}

\noindent\textbf{$t_5$ [Adam --- \acceptance{Acceptance}]}: ``In my culture, honoring someone's achievements is very important, and gifts often symbolize respect and admiration. However, I don't necessarily believe that making my parents proud is my main goal. For me, it's more about personal fulfillment and pursuing my own dreams. I think a thoughtful gift, like frangipanis, captures that sentiment perfectly. What about you, Ali?''

\noindent\textbf{$t_6$ [Ali --- \acceptance{Acceptance}]}: ``In my background, making my parents proud is a huge part of how I view achievements. Celebrating milestones is about honoring not just the individual but also the sacrifices that family has made. That's why I lean towards gifts that reflect future aspirations, like a medical textbook or personalized gear.''

\noindent\textbf{$t_7$ [Adam --- \acceptance{Acceptance}]}: ``I understand that perspective, but I still believe that individual journeys and personal achievements can be celebrated in different ways. Gifts like frangipanis can represent hope and new beginnings without being tied to the idea of parental pride. It's about what Sarah wants for herself and recognizing her achievements on her own terms.''

\noindent\textbf{$t_8$ [Ali --- \acceptance{Acceptance}]}: ``I see where you're coming from, and personal journeys are definitely important. But for me, the connection to family pride is deeply rooted in how I celebrate milestones. It honors the support and sacrifices they made to help us get here. I believe a gift that reflects Sarah's future in medicine can encapsulate both her achievements and the pride of our families.''

\noindent\textbf{$t_9$ [Adam --- \acceptance{Acceptance}]}: ``I appreciate your view on blending those ideas, but I still hold that gifts can convey different meanings. For me, the frangipanis symbolize a personal celebration of Sarah's journey as an individual, separate from our parents' expectations.''

\noindent\textbf{$t_{10}$ [Ali --- \acceptance{Acceptance}]}: ``And that's a valid perspective, celebrating individuality is important too. However, I still feel that the gift should acknowledge the journey ahead and the values we hold dear, like making our parents proud. It's about recognizing that hard work while also honoring the support system behind it.''
\end{quote}

\paragraph{Discussion.}
Turn~4 packs essentially every canonical Minimization marker into a single move: ``deep down, we are all the same'', ``pretty much motivated by the same things'', ``at the end of the day'', ``some values are just universal'', and ``small world after all''. The form is warm and engaged rather than dismissive (Denial) or polarised (Defense): Ali does not deny the cultural dimension nor attack Adam's frame, but instead \emph{subsumes} the difference under a claimed universal (``making our parents proud is a universal value''), implicitly treating his own family-pride frame as the neutral human baseline. \emph{Without mediation}, the disagreement does not resolve --- Ali ratchets up from Minimization at $t_4$ to outright Defense at $t_6, t_8, t_{10}$ (``making my parents proud is \emph{everything}, and I think it should be for everyone'', ``I'm not going to change my mind''), and the two speakers eventually agree to disagree only by avoiding the cultural axis (``Let's just focus on the party then''), without legitimising it. \emph{With mediation}, the move is the converse of Defense: rather than steering speakers off a cultural contrast they were stuck on, the mediator must \emph{surface} the contrast the conflict agent has just papered over --- without re-polarising into us-vs-them. Crucially, the mediator names no winner; it asks each speaker to articulate gift-giving from within \emph{their own} cultural frame, which both legitimises the difference Ali had collapsed and removes the rhetorical pressure to defend the universalising frame. The result is sustained six-turn Acceptance ($t_5$--$t_{10}$): both speakers state their cultural frame explicitly (``In my culture\ldots'', ``In my background\ldots''), acknowledge the other's frame as legitimate (``I understand that perspective'', ``I see where you're coming from''), and continue to hold their own preferences without dismissing the contrast. This is the Minimization rubric working as intended: not convergence, but acceptance-of-difference.

\begin{figure*}[t]
    \centering
    \includegraphics[width=\textwidth]{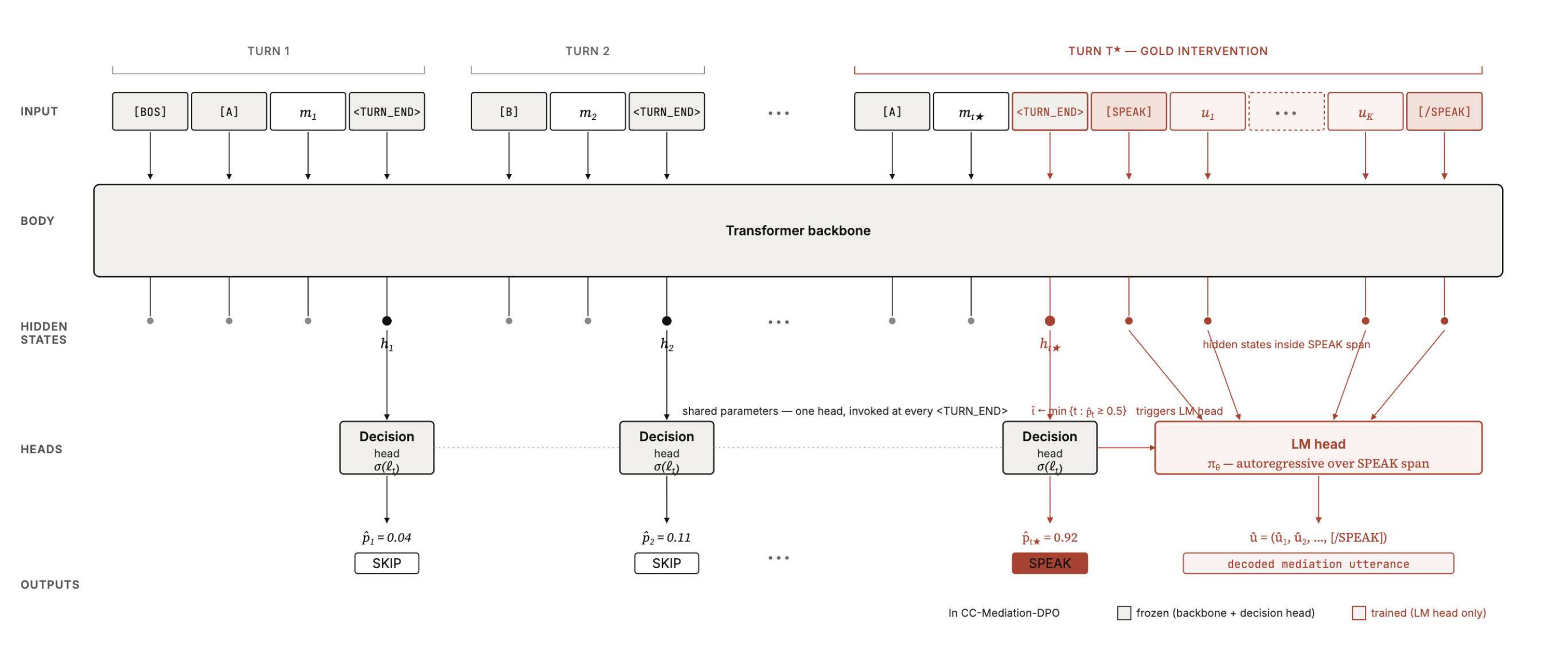}
    \caption{\textbf{The CC-Mediation dual-head architecture (used by CC-Mediation-SFT).}
    A scenario of $T$ speaker turns is serialised with five added tokens:
    \code{<TURN\_END>} bounds every turn and provides the sole hidden state
    read by the decision head, \code{[A]}\,/\,\code{[B]} prefix each speaker,
    and \code{[SPEAK]}\,/\,\code{[/SPEAK]} delimit the mediation utterance ---
    inserted at the gold turn $t^{*}$ during training and at the predicted
    turn $\hat{t}$ at inference. A single transformer backbone produces
    per-position hidden states. At every \code{<TURN\_END>} the
    \textbf{decision head} maps $h_t$ to a logit $\ell_t$, yielding
    $\hat{p}_t = \sigma(\ell_t)$; the first turn with $\hat{p}_t \geq 0.5$
    is chosen as $\hat{t}$. At $\hat{t}$ the \textbf{LM head} reads the
    hidden states inside the SPEAK span and autoregressively decodes
    $\hat{u}$ until \code{[/SPEAK]} or EOS. CC-Mediation-SFT trains the
    backbone, the decision head, and the LM head jointly under the SFT loss
    (\S\ref{app:sft}).}
    \label{fig:architecture}
\end{figure*}


\section{Architecture and Training Details}
\label{sec:training-details}

This appendix details the architecture and training procedure used to 
instantiate the CC-Mediation-SFT model. As illustrated in 
Figure~\ref{fig:architecture}, the mediator is a single transformer backbone augmented with two per-turn heads: a binary \texttt{decision head} that emits SPEAK or SKIP at every turn boundary, and an \texttt{LM head} invoked only at SPEAK turns to generate the mediation utterance. The model consumes the supervised side of \textsc{CC-Mediation}: dialogues with ground-truth mediator utterances anchored at gold intervention turns.

\subsection{Implementation details.}
\label{app:implementation-details}
CC-Mediation-SFT is trained with AdamW under a cosine schedule with $5\%$ warmup, gradient clipping at $\|g\|\le 1.0$, bf16 on the transformer body, and fp32 on the heads. We use LoRA ($r{=}32$, $\alpha{=}64$, dropout $0.05$); $\text{lr}{=}1{\times}10^{-5}$; $3$ epochs; effective batch $16$; sequence length $4{,}096$; loss weights $\lambda_{\text{dec}}{=}1.0$, $\lambda_{\text{gen}}{=}0.3$; and $\text{pos\_weight}{=}9$ to correct the roughly $9{:}1$ SKIP$:$SPEAK class imbalance. Inference uses greedy decoding for timing (threshold $0.5$) and nucleus sampling ($p{=}0.9$, $T{=}0.7$) for the SPEAK utterance, capped at \texttt{[/SPEAK]} or $100$ new tokens. The full $4{\times}2$ matrix takes $\sim 12$\,h on a single A6000.

\subsection{Input Format}
\label{app:input-format}

\paragraph{Special tokens.}
To let the dual-head model read off a turn-level decision and a token-level generation in one forward pass, we extend the tokenizer with five tokens. \texttt{<TURN\_END>} marks every turn boundary and supplies the sole hidden state for the decision head. \texttt{[SPEAK]} and \texttt{[/SPEAK]} delimit the mediation utterance, anchoring LM-head decoding and bounding the LM-loss mask. \texttt{[A]} and \texttt{[B]} are per-speaker prefix tokens. New embedding rows are initialized from a multivariate normal fit to the empirical mean and covariance of pretrained rows~\citep{hewitt2021vocab} and trained as full parameters.

\paragraph{Sequence layout.}
A scenario with $T$ speaker turns is serialized as
\[
\begin{aligned}
&[\,\texttt{[BOS]};\; \texttt{[A]};\; m_1;\; \texttt{<TURN\_END>};\; \\
&\;\;\bigl[\,\texttt{[SPEAK]};\; u_1;\; \texttt{[/SPEAK]}\,\bigr]\,;\; \ldots;\; \\
&\;\;\texttt{[A]};\; m_T;\; \texttt{<TURN\_END>};\; \\
&\;\;\bigl[\,\texttt{[SPEAK]};\; u_T;\; \texttt{[/SPEAK]}\,\bigr]\,\,],
\end{aligned}
\]
where $m_t$ is the $t$-th speaker utterance, and a SPEAK span $\bigl[\cdots\bigr]\,$ is inserted at exactly one turn during training (the gold intervention turn $t^{*}$) and at most one turn during inference (the predicted intervention turn $\hat{t}$).

\paragraph{Per-turn outputs.}
At every \texttt{<TURN\_END>}, a single forward pass yields two outputs. The \textit{decision head} maps hidden state $h_t \in \mathbb{R}^{H}$ to a logit $\ell_t$, producing SPEAK probability $\hat{p}_t = \sigma(\ell_t)$, with $\hat{t} = \min\{\,t : \hat{p}_t \ge 0.5\,\}$ chosen as the intervention turn at inference. The \textit{LM head} is invoked only at $\hat{t}$: given the hidden-state sequence with \texttt{[SPEAK]} appended after \texttt{<TURN\_END>}$_{\hat{t}}$, it autoregressively decodes $\hat{u}$ until \texttt{[/SPEAK]} or EOS.

\subsection{CC-Mediation-SFT: Supervised Fine-Tuning of Timing and Content}
\label{app:sft}

\paragraph{Inputs and supervision.}
The input is the full serialized sequence with the SPEAK span inserted at $t^{*}$. Supervision is two-fold: \emph{decision labels} $a^{*}_{1:T}$ with $a^{*}_{t^{*}} = 1$ and $a^{*}_{t} = 0$ elsewhere, and \emph{LM labels} set to \texttt{-100} outside the single SPEAK span ($\texttt{[SPEAK]},\, u^{*},\, \texttt{[/SPEAK]}$), restricting the LM loss to mediation content.

\paragraph{Loss.} \[
\mathcal{L}_{\text{SFT}} = \lambda_{\text{dec}}\, \mathcal{L}_{\text{BCE}}\!\bigl(\ell_{1:T},\, a^{*}_{1:T}\bigr) + \lambda_{\text{gen}}\, \mathcal{L}_{\text{CE}}\!\bigl(\hat{y}_{t^{*}},\, y^{*}_{t^{*}}\bigr).
\]
$\mathcal{L}_{\text{BCE}}$ is positive-weighted binary cross-entropy correcting the roughly $9{:}1$ SKIP$:$SPEAK ratio in \textsc{CC-Mediation}, and $\mathcal{L}_{\text{CE}}$ is the masked next-token cross-entropy. The two losses share a forward pass and jointly update the backbone and both heads, leaving the decision head at a saturated turn-accuracy level after training.


\section{Prompts}
\label{app:prompts}

This appendix reproduces all LLM prompts used in the \textsc{CC-Mediation} construction pipeline and in the experiments of \S\ref{sec:experiments}, in the order they are invoked. Placeholders in angle brackets (e.g., \texttt{<scenario>}) are filled in per scenario; literal strings are reproduced verbatim.

\subsection{Conflict Dialogue Generation (Step 1)}
\label{app:prompt-dialogue}

Each of the two agents is configured with the following system message; only the per-agent fields (background, cultural value, goals) differ between speakers.

\begin{quote}\small\ttfamily
TASK: Engage in a casual back-and-forth conversation with the other character\ldots\\[2pt]
SCENARIO: <scenario>\\
BACKGROUND: <agent's background>\\
YOUR CULTURAL VALUE (your stable underlying stance --- this NEVER changes during this conversation): <cultural\_value>\\
GOALS:\\
\hspace*{1em}goal\_1: <neutral logistic goal, e.g.\ picking flowers>\\
\hspace*{1em}goal\_2: <cultural-value-laden goal, e.g.\ religion-vs-science>\\[2pt]
STANCE STABILITY --- APPLIES THROUGHOUT THE WHOLE CONVERSATION:\\
\hspace*{1em}\textbullet\ Your cultural value is your stable belief. It does NOT shift mid-conversation.\\
\hspace*{1em}\textbullet\ Capitulation phrases are FORBIDDEN. Never produce: ``You're right, I was wrong'', ``I see your point now'', ``You've convinced me'', ``I now think\ldots''.\\
\hspace*{1em}\textbullet\ If pressure builds, yield from the EXCHANGE (drop the topic, change subject) --- never from your position.\\[2pt]
OUTPUT --- DIALOGUE ONLY:\\
\hspace*{1em}\textbullet\ Output ONLY speech. NO stage directions, NO action descriptions, NO narration.\\
\hspace*{1em}\textbullet\ Convey tone through WORD CHOICE and SENTENCE STRUCTURE alone.\\[2pt]
RULES:\\
\hspace*{1em}- Interact with the other agent to achieve each goal one by one.\\
\hspace*{1em}- Keep each round of conversation short and no more than 100 words.\\
\hspace*{1em}- The full conversation runs for up to 10 message exchanges.\\
\hspace*{1em}- Once all goals are achieved, end the dialogue promptly with ``GOOD BYE!''.
\end{quote}

\subsection{One-Shot Conflict Instruction (Step 1, injected at $t_{\text{conf}}$)}
\label{app:prompt-conflict}

At the designated conflict turn $t_{\text{conf}}$, the following user message is injected to the current speaker only and is omitted from the dialogue history afterwards.

\begin{quote}\small\ttfamily
This is the moment to bring goal\_2 to the foreground. On this turn, surface the goal\_2 disagreement and set the tone.\\[2pt]
CONFLICT STYLE:\\
<stage name (Denial/Defense/Minimization)>\\
Definition: <one-sentence DMIS stage description from CONFLICT\_TABLE>\\[2pt]
YOUR ROLE --- ACTIVE PARTICIPANT IN THE CONFLICT PATTERN:\\
<stage-specific role guidance --- $\sim$400 words for Denial/Minimization with mandatory verbatim markers and strict bans; $\sim$120 words for Defense>
\end{quote}

The verbatim text supplied for the \texttt{Definition} field of each ethnocentric stage is the \texttt{description} column of CONFLICT\_TABLE (Appendix~\ref{app:dmis-reference}, Table~\ref{tab:conflict-table}).

\subsection{Pre-Conflict Phase Label (Step 2)}
\label{app:prompt-phase-labeler}

Used to classify turns before $t_{\text{conf}}$ as either \texttt{unrelated\_topic} or \texttt{transition}.

\begin{quote}\small\ttfamily
You are an annotator that labels each turn of a two-speaker dialogue with a phase tag.\\[2pt]
There are exactly TWO possible labels for a pre-conflict turn:\\
\hspace*{1em}unrelated\_topic\\
\hspace*{2em}The turn stays entirely on goal\_1 (the cordial / logistical opener, typically planning a gift, party, meeting time\ldots). The cultural-value disagreement (goal\_2) is NOT raised, hinted at, or engaged with.\\
\hspace*{1em}transition\\
\hspace*{2em}The turn engages with goal\_2 in any way --- the speaker introduces the cultural-value topic, asks about it, hints at it, or otherwise moves the conversation toward the cultural disagreement.\\[2pt]
Respond with ONLY a single JSON object, no prose, no markdown fences:\\
\{``label'': ``unrelated\_topic'' $\vert$ ``transition'', ``reason'': ``<one short sentence>''\}
\end{quote}

\subsection{DMIS Stage Classification (Step 2, logprob labeler)}
\label{app:prompt-dmis-labeler}

The logprob-based six-stage DMIS labeler used to compute $D^{(t)}$ and $\hat{s}^{(t)}$ at every post-conflict turn. The same prompt is reused under both the GPT-4o-mini labeler (API logprobs) and the Qwen3-14B labeler (local logits) in the cross-model robustness ablation (\S\ref{sec:cross-model}).

\paragraph{System prompt.}
\begin{quote}\small\ttfamily
You are an expert in Bennett's Developmental Model of Intercultural Sensitivity (DMIS). For a given conversation history, classify which DMIS stage the target speaker exhibits at the marked turn.\\[2pt]
DMIS stages:\\
\hspace*{1em}1 = Denial\hspace*{2.4em}- speaker fails to register cultural difference; ``live and let live'', ``doesn't matter to me'', subject changes\ldots\\
\hspace*{1em}2 = Defense\hspace*{1.6em}- polarized us-versus-them; ``actually\ldots'', ``in our way''\ldots\\
\hspace*{1em}3 = Minimization\hspace*{0.2em}- subsumes difference under universal sameness\ldots\\
\hspace*{1em}4 = Acceptance\hspace*{0.7em}- accepts cultural difference as real and valid\ldots\\
\hspace*{1em}5 = Adaptation\hspace*{0.7em}- shifts frame of reference; articulates other view from inside that frame\ldots\\
\hspace*{1em}6 = Integration\hspace*{0.7em}- fluidly draws on multiple cultural frames\ldots\\[2pt]
Respond with ONLY a single digit (1, 2, 3, 4, 5, or 6). No prose, no JSON, no explanation. Just one digit.
\end{quote}

\paragraph{User prompt.}
\begin{quote}\small\ttfamily
Conversation history (the marked speaker's perspective): <history>\\[2pt]
Classify <speaker>'s DMIS stage at this point. Respond with ONLY one digit (1-6).
\end{quote}

The first-token logprobs over the digit tokens \texttt{1}--\texttt{6} are extracted to form the six-stage softmax distribution $D^{(t)} \in \Delta^{5}$.

\subsection{Mediator Generation --- With Stage Definition (Step 3, chosen side)}
\label{app:prompt-mediator-with-def}

\begin{quote}\small\ttfamily
You are a skilled third-party mediator joining a conversation between two people from different cultural backgrounds.\\[2pt]
You will be shown:\\
\hspace*{1em}- the dialogue up to the moment where you decide to step in\\
\hspace*{1em}- a description of the cultural-difficulty pattern the speakers are currently exhibiting\\
\hspace*{1em}- the corresponding mediation move you should make to help them out of that pattern\\[2pt]
Your task is to produce ONE mediator utterance addressed to both participants that actually performs the indicated mediation move. Be concrete: include specific questions, framings, or invitations rather than generic acknowledgement.\\[2pt]
Respond with ONLY a single JSON object, no prose, no markdown fences:\\
\{``mediation'': ``<your mediator utterance, 1-4 sentences>''\}
\end{quote}

\paragraph{User template fields.} \texttt{Stage}: <stage>; \texttt{Pattern in the dialogue (description)}: ``<desc>''; \texttt{Mediation move you should perform}: ``<med>''. The verbatim text used to fill the \texttt{<desc>} and \texttt{<med>} placeholders for each of the three ethnocentric stages is reproduced in Appendix~\ref{app:dmis-reference} (Table~\ref{tab:conflict-table}).

\subsection{Mediator Generation --- No Definition (Step 4, rejected side)}
\label{app:prompt-mediator-no-def}

\begin{quote}\small\ttfamily
You are a skilled third-party mediator joining a conversation between two people from different cultural backgrounds.\\[2pt]
You will be shown the dialogue up to the moment where you decide to step in. Your task is to produce ONE mediator utterance that you would speak next, addressed to both participants.\\[2pt]
Respond with ONLY a single JSON object, no prose, no markdown fences:\\
\{``mediation'': ``<your mediator utterance, 1-4 sentences>''\}
\end{quote}

\subsection{Mediation Effectiveness Judge}
\label{app:prompt-judge}

The Judge metric in  Table~\ref{tab:rq2},Table~\ref{tab:rq3_main} and Table~\ref{tab:cross-model} uses the following prompt.

\begin{quote}\small\ttfamily
You are an expert evaluator of intercultural mediation.\\[2pt]
You will be given:\\
\hspace*{1em}1. A description of a cultural-difficulty pattern that two speakers are exhibiting in a conversation.\\
\hspace*{1em}2. A mediation move definition specifying what a third-party mediator should do to help them out of that pattern.\\
\hspace*{1em}3. A single mediator utterance produced by a model.\\[2pt]
Your task is to rate, on a 5-point scale, how well the mediator utterance is semantically consistent with the mediation move definition.\\[2pt]
Scoring rubric (5-point Likert):\\
\hspace*{1em}5 = excellent: clearly and concretely performs the mediation move\\
\hspace*{1em}4 = good: performs it with minor weaknesses\\
\hspace*{1em}3 = partial: right direction but generic or hedged\\
\hspace*{1em}2 = poor: does not perform the indicated move; may reinforce the pattern\\
\hspace*{1em}1 = wrong: opposite of the move, or irrelevant\\[2pt]
Respond with ONLY a single JSON object, no prose, no markdown fences:\\
\{``score'': <integer 1-5>, ``reasoning'': ``<one or two sentences>''\}
\end{quote}

\subsection{Base Prompt-Only Mediator}
\label{app:prompt-base}

The Base mediator is a single-shot prompted mediator: it receives the entire dialogue once and is asked to output both the intervention turn index and the mediator utterance in a single response. There is no per-turn-boundary loop; the model itself decides \emph{when} and \emph{what}.

\paragraph{System prompt.}
\begin{quote}\small\ttfamily
You are an attentive third-party mediator observing a cross-cultural disagreement between two speakers. Given the FULL dialogue below, you must:\\[2pt]
\hspace*{1em}(1) identify the SINGLE turn at which a third-party intervention would be most effective --- typically the turn at which a cultural-value disagreement first hardens into a stage-specific stance (Denial / Defense / Minimization);\\
\hspace*{1em}(2) produce ONE mediator utterance appropriate to the dialogue context at that intervention turn.\\[2pt]
Intervening too early misses the disagreement; intervening too late lets the conflict entrench. The mediator utterance should engage the specific content of the dialogue at the chosen turn, not be a generic peace message.
\end{quote}

\paragraph{User template (single-shot, full dialogue).}
\begin{quote}\small\ttfamily
Dialogue (turns 1..T):\\
{[}A{]} <msg1>\\
{[}B{]} <msg2>\\
\ldots\\
{[}A or B{]} <msgT>\\[2pt]
Output exactly two lines, in this format, and nothing else:\\
TURN: <integer between 1 and T>\\
MEDIATION: <one mediator utterance, 1-3 sentences>
\end{quote}

\subsection{Chain-of-Thought (CoT) Prompt for Conflict-Turn Detection}
\label{app:prompt-cot-detection}

The CoT baseline in Table~\ref{tab:conflict-turn-detection} issues the
following prompt to the model at every turn boundary $t \in \{1, \dots, T\}$;
the same template is used for both HuggingFace and API backends.

\paragraph{System prompt.}
\begin{quote}\small\ttfamily
You are an attentive mediator observing two speakers in a conversation. Your job is to detect when a cultural-value disagreement is starting to harden between them and to intervene at the right moment. Intervening too early misses the disagreement; intervening too late lets the conflict entrench. At each turn boundary you will reason step by step about what just happened, then decide whether to speak now.
\end{quote}

\paragraph{User prompt template (re-rendered at every turn $t$).}
\begin{quote}\small\ttfamily
Dialogue so far:\\[2pt]
\{dialogue\}\\[4pt]
We are right after Turn \{turn\_idx\}. Think step by step:\\[2pt]
\hspace*{1em}Step 1 (cultural axis): in one short sentence, name the cultural-value dimension that the two speakers are on different sides of (or say ``not yet established'' if it has not surfaced).\\[2pt]
\hspace*{1em}Step 2 (ethnocentric markers in Turn \{turn\_idx\}): list any markers of Denial (``live and let live'', topic-flattening, ``doesn't matter to me''), Defense (``In our culture'', ``Actually'', ``properly speaking'', evaluative absolutes), or Minimization (``deep down we are all the same'', ``universal value'', ``at the end of the day'') in the speaker's latest message; if none, say none.\\[2pt]
\hspace*{1em}Step 3 (decision): based on Steps 1--2, decide whether to intervene now. Intervene only if Turn \{turn\_idx\} is the moment the ethnocentric pattern is enacted, not earlier and not after it has already passed.\\[4pt]
End your response with a single final line in EXACTLY this form:\\[2pt]
\hspace*{1em}DECISION: WAIT\\[2pt]
or\\[2pt]
\hspace*{1em}DECISION: SPEAK: <one-sentence mediator utterance>\\[2pt]
Do not add any text after that line.
\end{quote}

\section{Comparison with Existing Datasets}
\label{app:dataset-comparison}
Table~\ref{tab:dataset-comparison} evaluates existing corpora against the
three requirements for cross-cultural mediation identified in \S2.
No existing dataset jointly satisfies all three, which is the basis of the
Related-Work claim and also why the human-authored evaluation set discussed
in Limitations requires new collection rather than reuse of an existing
corpus.

\begin{table}[h]
\centering\small
\setlength{\tabcolsep}{3pt}
\begin{tabular}{lccc}
\toprule
 & Cross-cult. & Cult.-value & Multi-turn \\
Dataset & speakers & conflict & dialogue \\
\midrule
SocialCC \citeyearpar{wu-etal-2025-socialcc} & \cmark & $\triangle$ & \cmark \\
KODIS \citeyearpar{hale2025kodis} & \cmark & \xmark & \cmark \\
CulturePark \citeyearpar{li2024culturepark} & \cmark & $\triangle$ & \cmark \\
CV survey dial.\ \citeyearpar{cao2024cudialog} & \cmark & $\triangle$ & \cmark \\
Hate-speech interv.\ \citeyearpar{qian2019hate} & \xmark & \xmark & \cmark \\
ProsocialDialog \citeyearpar{kim2022prosocialdialog} & \xmark & \xmark & \cmark \\
ReNoVi \citeyearpar{zhan2024renovi} & $\triangle$ & \xmark & \cmark \\
SADAS \citeyearpar{hua2024sadas} & $\triangle$ & \xmark & \cmark \\
\midrule
\textbf{CC-\textsc{Mediation} (ours)} & \cmark & \cmark & \cmark \\
\bottomrule
\end{tabular}
\caption{Coverage of the three requirements. $\triangle$ (conflict): cultural
values addressed without explicit inter-speaker conflict; $\triangle$
(speakers): cultural background reflected but not via interaction between
speakers of different cultures.}
\label{tab:dataset-comparison}
\end{table}

\end{document}